\documentclass[10pt,twocolumn,letterpaper]{article}

\usepackage[final]{iccv} 
\usepackage{multirow}
\usepackage{graphicx}
\usepackage{color}
\usepackage{float,eucal}
\definecolor{cvprblue}{rgb}{0.21,0.49,0.74}
\usepackage[pagebackref,breaklinks,colorlinks,allcolors=cvprblue]{hyperref}

\def\paperID{3071} % *** Enter the Paper ID here
\def\confName{ICCV}
\def\confYear{2025}

\newcommand{\figref}[1]{Fig.~\ref{#1}}
\newcommand{\tabref}[1]{Table~\ref{#1}}
\newcommand{\equref}[1]{Eqn.~\eqref{#1}}
\newcommand{\secref}[1]{Sec.~\ref{#1}}
\newcommand{\name}{SurfaceSplat\xspace}

\makeatletter
\newcommand{\printfnsymbol}[1]{%
  \textsuperscript{\@fnsymbol{#1}}%
}
\makeatother

\begin{document}

% %%%%%%%%% TITLE - PLEASE UPDATE
% % \title{Enhanced Sparse View Surface Reconstruction via Novel View Rendering}
\title{SurfaceSplat: Connecting Surface Reconstruction and Gaussian Splatting}

%%%%%%%%% AUTHORS - PLEASE UPDATE
% \author{Zihui Gao$^{*1}$\\
% % {\tt\small firstauthor@i1.org}
% \and
% Jia-Wang Bian
% }

% \author{Zihui Gao\thanks{Equal contributions} \inst{1} 
% \and
% Jia-Wang Bian \printfnsymbol{1}
% \inst{2} 
% \and
% Guosheng Lin\inst{3}
% \and Hao Chen\inst{1}
% \and Chunhua Shen\inst{1}
% }
\author{
Zihui Gao\textsuperscript{1,3}\thanks{Equal contributions},
~~ 
Jia-Wang Bian\textsuperscript{2}\footnotemark[1], 
~~
Guosheng Lin\textsuperscript{3},
~~
Hao Chen\textsuperscript{1}\thanks{Corresponding author},
~~
Chunhua Shen\textsuperscript{1} \\[.2cm]
\textsuperscript{1}Zhejiang University, China \quad
\textsuperscript{2}ByteDance Seed \quad
\textsuperscript{3}Nanyang Technological University, Singapore \\
\small
\textsuperscript{*}Equal contribution \quad
\textsuperscript{\dag}Corresponding author
}

\twocolumn[{%
\renewcommand\twocolumn[1][]{#1}%
\maketitle
\begin{center}
    \centering
\includegraphics[width=1.0\linewidth]{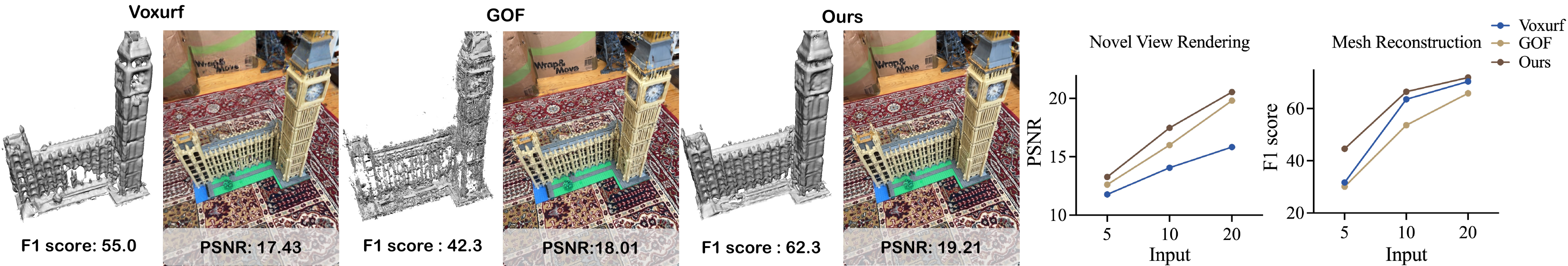}
% {figure/teaser_big_ben.pdf}
\captionof{figure}{\textbf{Sparse view reconstruction and rendering comparison}.
\textit{Left:} Qualitative results from 10 images evenly sampled from a casually captured 360-degree video. 
% results are generated from 10 images evenly sampled from a casually captured 360-degree video. We observe that SDF-based methods (e.g., Voxurf) excel at global geometry but struggle with fine details due to their dense nature, while 3DGS-based methods (e.g., 2DGS and GOF) capture local details well but lack global coherence due to their sparse nature. Leveraging these complementary properties, we propose \name, a hybrid approach that bridges both representations for optimal performance, leading to significant improvements in both surface reconstruction and novel view synthesis.
\textit{Right:}
Quantitative analysis of 5, 10, and 20 input views, averaged across the selected 9 MobileBrick test scenes.
  3DGS-based methods (e.g., GOF) achieve superior novel view rendering than SDF-based methods (e.g., Voxurf) due to their sparse representations, which capture fine details. However, SDF-based methods outperform the former in mesh reconstruction, as their dense representations better preserve global geometry. Our approach combines the strengths of both, achieving optimal performance.
}
\label{fig:teaser}
\end{center}
}]

\begin{abstract}
Surface reconstruction and novel view rendering from sparse-view images are challenging. 
Signed Distance Function (SDF)-based methods struggle with fine details, while 3D Gaussian Splatting (3DGS)-based approaches lack global geometry coherence. 
We propose a novel hybrid method that combines the strengths of both approaches: SDF captures coarse geometry to enhance 3DGS-based rendering, while newly rendered images from 3DGS refine the details of SDF for accurate surface reconstruction. As a result, our method surpasses state-of-the-art approaches in surface reconstruction and novel view synthesis on the DTU and MobileBrick datasets. Code will
be released at:
\href{https://github.com/aim-uofa/SurfaceSplat}{https://github.com/aim-uofa/SurfaceSplat}.

% Surface reconstruction from sparse viewpoint images poses significant challenges. Signed Distance Function (SDF)-based approaches often struggle to capture fine details, while 3D Gaussian Splatting (3DGS)-based methods frequently produce discontinued surfaces. To achieve high-quality reconstruction in sparse-view settings, we propose a novel approach that combines the strengths of both worlds. Our method first constructs a coarse mesh using SDF-based representations. Next, it samples point clouds from the object surface to initialize the 3DGS-based method, and finally, we rendered novel view images under expanded poses via 3DGS to refine the coarse mesh. Comprehensive experiments and ablation studies on two real-world datasets, DTU and MobileBrick, demonstrate that our method outperforms previous state-of-the-art approaches in both surface reconstruction and novel view synthesis. 
% The code will be released upon acceptance.
\end{abstract}    
\section{Introduction}
\label{sec:intro}

% \begin{figure}
%   \centering
%   \includegraphics[width = 0.5\textwidth]{figure/intro.pdf}
%   \caption{\textcolor{red}{occupy the intro position}.}
%   \label{fig:intro}
% \end{figure}

% 从多张图像重建出物体三维表征在电商，虚拟现实，机器人，自动驾驶等领域有着重要的作用。（）

% 三维表征是多样的，物体几何表面mesh，occupancy以及近期热点3DGS。现在已经有很多表现很好的算法能实现dense image 的三维重建。
% 讨论一下相关工作 1，2句话
% 3D reconstruction from multi-view images is a fundamental problem in computer vision, with applications spanning virtual reality, robotics, and autonomous driving. Recent advancements in Neural Radiance Fields (NeRF)~\cite{mildenhall2020nerf} and 3D Gaussian Splatting (3DGS)~\cite{kerbl20233d} have significantly pushed the field forward. However, their performance degrades substantially with limited viewpoint images—a common limitation in real-world scenarios. This paper addresses sparse-view reconstruction, aiming to enhance surface reconstruction and novel view synthesis.

3D reconstruction from multi-view images is a core problem in computer vision with applications in virtual reality, robotics, and autonomous driving. Recent advances in Neural Radiance Fields (NeRF)~\cite{mildenhall2020nerf} and 3D Gaussian Splatting (3DGS)~\cite{kerbl20233d} have significantly advanced the field. However, their performance degrades under sparse-view conditions, a common real-world challenge. This paper tackles sparse-view reconstruction to bridge this gap.
Unlike approaches that leverage generative models~\cite{xu2024instantmesh,wu2024direct3d,chen2024meshanything,wei2024meshlrm} or learn geometry priors through large-scale pretraining~\cite{yu2021pixelnerf,chen2021mvsnerf,long2022sparseneus,ren2023volrecon}, we focus on identifying the optimal 3D representations for surface reconstruction and novel view synthesis.

% Recent methods for surface reconstruction primarily rely on either continuous SDF-based or 3D Gaussian Splatting (3DGS)-based representations. SDF-based methods, such as NeuS~\cite{wang2021neus} and Voxurf~\cite{wu2022voxurf}, use Multi-Layer Perceptrons (MLPs) or voxel grids to model scene geometry, leveraging differentiable volume rendering~\cite{mildenhall2020nerf} to synthesize novel view images and optimize geometry by comparing rendered and ground-truth views. In contrast, 3DGS-based methods like Gaussian Opacity Field (GOF)\cite{Yu2024GOF} and 2DGS\cite{Huang2DGS2024} employ differentiable rasterization to render images, refining color and shape through pre-computed point clouds and progressively densifying points. However, both SDF- and 3DGS-based methods struggle with poor reconstruction quality under sparse viewpoint conditions due to insufficient regularization, as illustrated in \figref{fig:teaser}.

Surface reconstruction methods primarily use the Signed Distance Function (SDF) or 3DGS-based representations.
Here, SDF-based approaches, such as NeuS~\cite{wang2021neus} and Voxurf~\cite{wu2022voxurf}, model scene geometry continuously with dense representations and optimize them via differentiable volume rendering~\cite{mildenhall2020nerf}. In contrast, 3DGS-based methods like GOF~\cite{Yu2024GOF} and 2DGS~\cite{Huang2DGS2024} leverage a pre-computed sparse point cloud for image rendering and progressively densify and refine it through differentiable rasterization.
Due to their dense representations, SDF-based methods capture global structures well but lack fine details, while the sparse nature of 3DGS-based methods enables high-frequency detail preservation but compromises global coherence.
As a result, both approaches struggle with poor reconstruction quality under sparse-view conditions. Typically, SDF-based methods outperform 3DGS in surface reconstruction, while 3DGS excels in image rendering, as illustrated in \figref{fig:teaser}.

% \figref{fig:teaser} \textit{left} illustrates visual results. 
% In \figref{fig:teaser} \textit{right}, we present results across different levels of sparsity, where Voxurf consistently demonstrates superior surface reconstruction capability, while GOF excels in rendering quality. Therefore, we integrate their strengths.

% To better understand their nature, we first conduct an experimental exploration of the performance of mainstream methods under varying levels of sparsity. Detailed results are provided in the supplementary materials.
% We observe that SDF-based methods~\cite{Yu2022MonoSDF,wu2022voxurf} struggle to capture fine details in sparser settings, as they rely on the overlap of densely sampled images to optimize a single MLP function. A similar trend is observed in \figref{fig:teaser} and ~\figref{fig:mesh_vis}. While 3DGS-based methods have difficulty reconstructing complete global geometry due to the excessively sparse initialization of point clouds.

\begin{figure*}
  \centering
  \includegraphics[width = 0.905\textwidth]{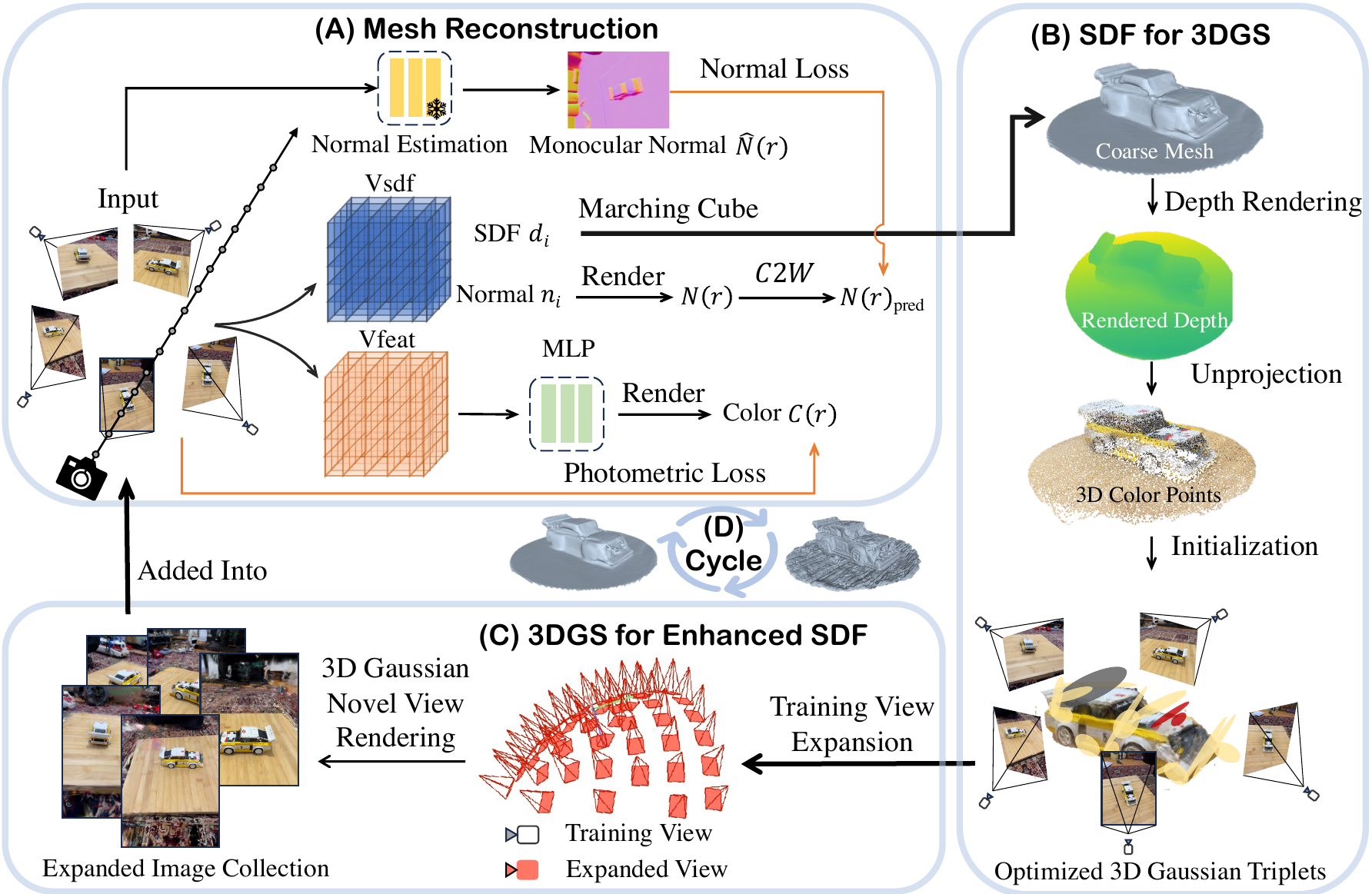}
  \caption{Overview of the proposed \name. (A) We reconstruct a coarse mesh using an SDF-based representation. (B) Point clouds are sampled from the mesh surface to initialize 3DGS. (C) 3DGS renders new viewpoint images to expand the training set, refining the mesh. (D) Steps B and C can be repeated for iterative optimization, progressively improving performance.}
  \label{fig:method}
\end{figure*}

Recognizing the complementary strengths of SDF-based (dense) and 3DGS-based (sparse) representations, we propose a novel hybrid approach, \name, as illustrated in \figref{fig:method}. Our method is built on two key ideas:
(i) \textbf{SDF for Improved 3DGS}: To address the limitation of 3DGS in learning global geometry, we first fit the global structure using an SDF-based representation, rapidly generating a smooth yet coarse mesh. We then initialize 3DGS by sampling point clouds from the mesh surface, ensuring global consistency while allowing 3DGS to refine fine details during training.
(ii) \textbf{3DGS for Enhanced SDF}: To compensate for the inability of SDF-based methods to capture fine details under sparse-view settings, we leverage the improved 3DGS from the first step to render additional novel viewpoint images, expanding the dataset. This enriched supervision helps the SDF-based method learn finer structural details, leading to improved reconstruction quality.

We conduct experiments on two real-world datasets, DTU~\cite{DTU2014} and MobileBrick~\cite{li2023mobilebrick}.
Our method, \name, achieves state-of-the-art performance in sparse-view novel view rendering and 3D mesh reconstruction.
In summary, we make the following contributions:

\begin{itemize}
    \item We propose \name, which synergistically combines the strengths of SDF-based and 3DGS-based representations to achieve optimal global geometry preservation while capturing fine local details.

    % \item We propose a novel mutual-help framework, in which we enhance 3DGS by sampling points from the mesh surface,
    % and we improve mesh reconstruction by using newly rendered images with our expansion camera poses.

    \item We conducted a comprehensive evaluation and ablations on DTU and MobileBrick datasets. \name achieves state-of-the-art performance in novel view synthesis and mesh reconstruction under sparse-view conditions.
    
\end{itemize}

% \begin{figure}
%   \centering
%   \includegraphics[width = 0.47\textwidth]{figure/Teaser_sparisty.png}
%   \caption{Quantitative analysis on MobileBrick.
%   3DGS-based methods (e.g., GOF) achieve superior novel view rendering than SDF-based methods (e.g., Voxurf) due to their sparse representations, which capture fine details. However, SDF-based methods outperform the former in mesh reconstruction, as their dense representations better preserve global geometry. Our approach combines the strengths of both, achieving optimal performance.}
%   \label{fig:3dgs_psnr}
% \end{figure}

\section{Related work}
\label{sec:related_work}

%-------------------------------------------------------------------------
\subsection{Novel View Synthesis from Sparse Inputs}
% Neural Radiance Fields (NeRFs)~\cite{mildenhall2020nerf} introduced implicit 3D representation for enhanced view rendering quality but suffer from long training time. NeRF’s success spurred numerous methods~\cite{barron2021mip,barron2022mip,zhang2020nerf++} leverage regularization strategies to improve its quality and speed. Instant-NGP~\cite{muller2022instant} uses a hash and occupancy grid for acceleration, along with a compact MLP for density and appearance, while DVGO~\cite{sun2022direct} combines voxel grids with a shallow MLP in a hybrid design to accelerate. 
% Additionally, the recently developed 3D Gaussian Splatting (3DGS)~\cite{kerbl20233d} achieves efficient training and real-time rendering. Instead of fitting a neural network, it optimizes parameters of a set of 3D ellipsoids, including position, opacity, scale, rotation, and spherical harmonics. The 3D Gaussians are splatted onto the image plane and rendered via alpha blending.
% When dealing with sparse inputs, SparseGS~\cite{xiong2023sparsegs} proposes using enhanced 3D Gaussians for novel view synthesis, incorporating a depth correction loss and a diffusion model to refine the output. However, the quality of 3DGS-based work is often affected by the initial points~\cite{jung2024relaxing}, especially when sparse views yield extremely sparse and uneven SfM points. Our proposed strategy of uniform sampling on the mesh for initializing 3D Gaussians significantly enhances 3DGS rendering quality under sparse view conditions.

Neural Radiance Fields (NeRFs)-based methods~\cite{mildenhall2020nerf,barron2021mip,barron2022mip,zhang2020nerf++,chen2022tensorf,fridovich2022plenoxels,muller2022instant,sun2022direct,turkulainen2024dn,yang2024deformable,bian2022nopenerf} have revolutionized novel view synthesis with implicit neural representations, and 3DGS-based methods~\cite{kerbl20233d, yu2024mip, lu2024scaffold, ye2024gsplat, song2024sa, sun20243dgs, yang2024spectrally} enable efficient training and real-time rendering through explicit 3D point clouds. However, both approaches suffer from performance degradation in sparse-view settings.
To address this issue, recent methods have explored generative models~\cite{xu2024instantmesh,wu2024direct3d,chen2024meshanything,wei2024meshlrm} or leveraged large-scale training to learn geometric priors~\cite{yu2021pixelnerf,chen2021mvsnerf,long2022sparseneus,ren2023volrecon}. Unlike these approaches, we argue that the key challenge lies in the lack of effective geometric initialization for 3DGS. To overcome this, we investigate how neural surface reconstruction methods can enhance its performance.

% SparseGS\cite{xiong2023sparsegs} builds on this by refining view synthesis with depth correction and diffusion models for sparse inputs.

% However, 3DGS methods often struggle with uneven Structure-from-Motion (SfM) points under sparse views~\cite{jung2024relaxing}. Our method improves 3DGS by incorporating uniform mesh sampling for initialization, significantly enhancing rendering quality in sparse settings.

%-------------------------------------------------------------------------
\subsection{Neural Surface Reconstruction}

SDF-based methods, such as NeuS~\cite{wang2021neus}, VolSDF~\cite{yariv2021volsdf}, Neuralangelo~\cite{li2023neuralangelo}, and PoRF~\cite{bian2024porf} use dense neural representations and differentiable volume rendering to achieve high-quality reconstructions with 3D supervision. However, they suffer from long optimization times and require dense viewpoint images.
Recent methods, such as 2DGS~\cite{Huang2DGS2024} and GOF~\cite{Yu2024GOF}, extend 3DGS~\cite{kerbl20233d} by leveraging modified Gaussians and depth correction to accelerate geometry extraction. While 3DGS-based methods~\cite{guedon2024sugar, Huang2DGS2024, Yu2024GOF,zhang2024rade,chen2024pgsr,kim2024integrating,wang2024enhancement,bai2024360} excel at capturing fine local details, their sparse representations struggle to maintain global geometry, leading to incomplete and fragmented reconstructions.
This paper focuses on integrating the strengths of both representations to achieve optimal neural surface reconstruction.

%-------------------------------------------------------------------------
\subsection{Combing 3DGS and SDF}
Several recent approaches have integrated SDF-based~\cite{osher2003constructing, park2019deepsdf} and 3DGS-based representations to improve surface reconstruction. NeuSG~\cite{chen2023neusg} and GSDF~\cite{yu2024gsdf} jointly optimize SDF and 3DGS, enforcing geometric consistency (e.g., depths and normals) to improve surface detail~\cite{he2024nerfsback}. Similarly, 3DGSR~\cite{lyu20243dgsr} combines SDF values with Gaussian opacity in a joint optimization framework for better geometry. While effective in dense-view settings, these methods struggle to reconstruct high-quality structures under sparse-view conditions, as shown in our experiments in ~\secref{sec:experiment}. Our approach specifically targets sparse-view scenarios by leveraging a complementary structure to enhance both rendering and reconstruction quality.

\section{Method}
\label{sec:method}
Our method takes sparse viewpoint images with camera poses as input, aiming to reconstruct 3D geometry and color for novel view synthesis and mesh extraction. \figref{fig:method} provides an overview of \name.
In the following sections, we first introduce the preliminaries in \secref{method:preliminaries}, then explain how SDF-based mesh reconstruction improves 3DGS for novel view synthesis in \secref{method:meshing}, and finally describe how 3DGS-based rendering enhances SDF-based surface reconstruction quality in \secref{method:rendering}.

\subsection{Preliminaries}\label{method:preliminaries}

\textbf{SDF-based representation.}
% DVGO
% voxurf
NeuS~\cite{wang2021neus} proposes to model scene coordinates as signed distance function (SDF) values and optimize using differentiable volume rendering, similar to NeRF~\cite{mildenhall2020nerf}. After optimization, object surfaces are extracted using the marching cubes algorithm~\cite{lorensen1998marching}. To render a pixel, a ray is cast from the camera center $o$ through the pixel along the viewing direction $v$ as $\{p(t) = o + tv | t \geq 0\}$, and the pixel color is computed by integrating $N$ sampled points along the ray $\{p_i = o + t_iv | i=1, ..., N, t_i < t_{i+1} \}$ using volume rendering:
\begin{equation}
\hat{C}(r) = \sum_{i=1}^{N}T_i \alpha_i c_i, \ \ T_i = \prod \limits_{j=1}^{i-1} (1-\alpha_j),
\end{equation}
where $\alpha_i$ represents opacity and $T_i$ is the accumulated transmittance.
It is computed as:
\begin{equation}
\alpha_i = \max\left(\frac{\Phi_s(f(p(t_i))) - \Phi_s(f(p(t_{i+1})))}{\Phi_s(f(p(t_i)))}, 0\right),
\end{equation}
where $f(x)$ is the SDF function and $\Phi_s(x) = (1 + e^{-sx})^{-1}$ is the Sigmoid function, with $s$ learned during training. 
Based on this, Voxurf~\cite{wu2022voxurf} proposes a hybrid representation that combines a voxel grid with a shallow MLP to reconstruct the implicit SDF field. In the coarse stage, Voxurf~\cite{wu2022voxurf} optimizes for a better overall shape by using 3D convolution and interpolation to estimate SDF values.
In the fine stage, it increases the voxel grid resolution and employs a dual-color MLP architecture, consisting of two networks: $g_{geo}$, which takes hierarchical geometry features as input, and 
$g_{feat}$, which receives local features from 
$\boldsymbol{V}^{(\mathrm{feat})}$ along with surface normals. We incorporate Voxurf in this work due to its effective balance between accuracy and efficiency.

% Direct Voxel Grid Optimization (DVGO) introduces a voxel grid representation that explicitly models properties of interest (such as density, color, or features) within its grid cells. This explicit scene representation allows efficient querying of any 3D position via interpolation. Specifically, it employs a dense voxel grid $\boldsymbol{V}^{(\mathrm{density})} \in \mathbb{R}^{1 \times N_x \times N_y \times N_z}$ to represent the scene geometry and the raw volume density is 
% \[
% \sigma=\operatorname{interp}\left(\boldsymbol{x}, \boldsymbol{V}^{(\mathrm{density})}\right)
% \]
% complemented by a feature grid $\boldsymbol{V}^{(\mathrm{feat})} \in \mathbb{R}^{C \times N_x \times N_y \times N_z}$ combined with a shallow MLP to model the appearance, as follow:
% \[
% \boldsymbol{c}=\operatorname{MLP}_{\Theta}\left(\operatorname{interp}\left(\boldsymbol{x}, \boldsymbol{V}^{(\mathrm{feat})}\right), \boldsymbol{x}, \boldsymbol{d}\right)
% \]
% where the $\boldsymbol{x}$ and $\boldsymbol{d}$ are 3D query points and directions.

% Voxel-based surface
% reconstruction(Voxurf), to achieve more accurate surface reconstruction, employs $\boldsymbol{V}^{(\mathrm{sdf})}$ and 
\paragraph{3DGS-based representation.}
3DGS~\cite{kerbl20233d} models a set of 3D Gaussians to represent the scene, which is similar to point clouds. Each Gaussian ellipse has a color and an opacity and is defined by its centered position $x$ (mean), and a full covariance matrix $\Sigma$:
% \begin{equation}
$G(x) = e^{-\frac{1}{2} x^T \Sigma^{-1} x}$.
% \end{equation}
When projecting 3D Gaussians to 2D for rendering, the splatting method is used to position the Gaussians on 2D planes, which involves a new covariance matrix $\Sigma'$ in camera coordinates defined as:
% \begin{equation}
$\Sigma' = J W \Sigma W^T J^T$,
% \end{equation}
where $W$ denotes a given viewing transformation matrix and $J$ is the Jacobian of the affine approximation of the projective transformation. To enable differentiable optimization, $\Sigma$ is further decomposed into a scaling matrix $S$ and a rotation matrix $R$:
% \begin{equation}
$\Sigma = RSS^T R^T$.
% \end{equation}

% 稀疏时 colmap初始化点云差，注意这里为了重建结果和gt进行评测，输入图像的gt pose是需要的，然后进行有gt pose的colmap初始化得到点云。

\subsection{SDF for Improved 3DGS}\label{method:meshing}
\begin{figure}
  \centering
  \includegraphics[width = 0.48\textwidth]{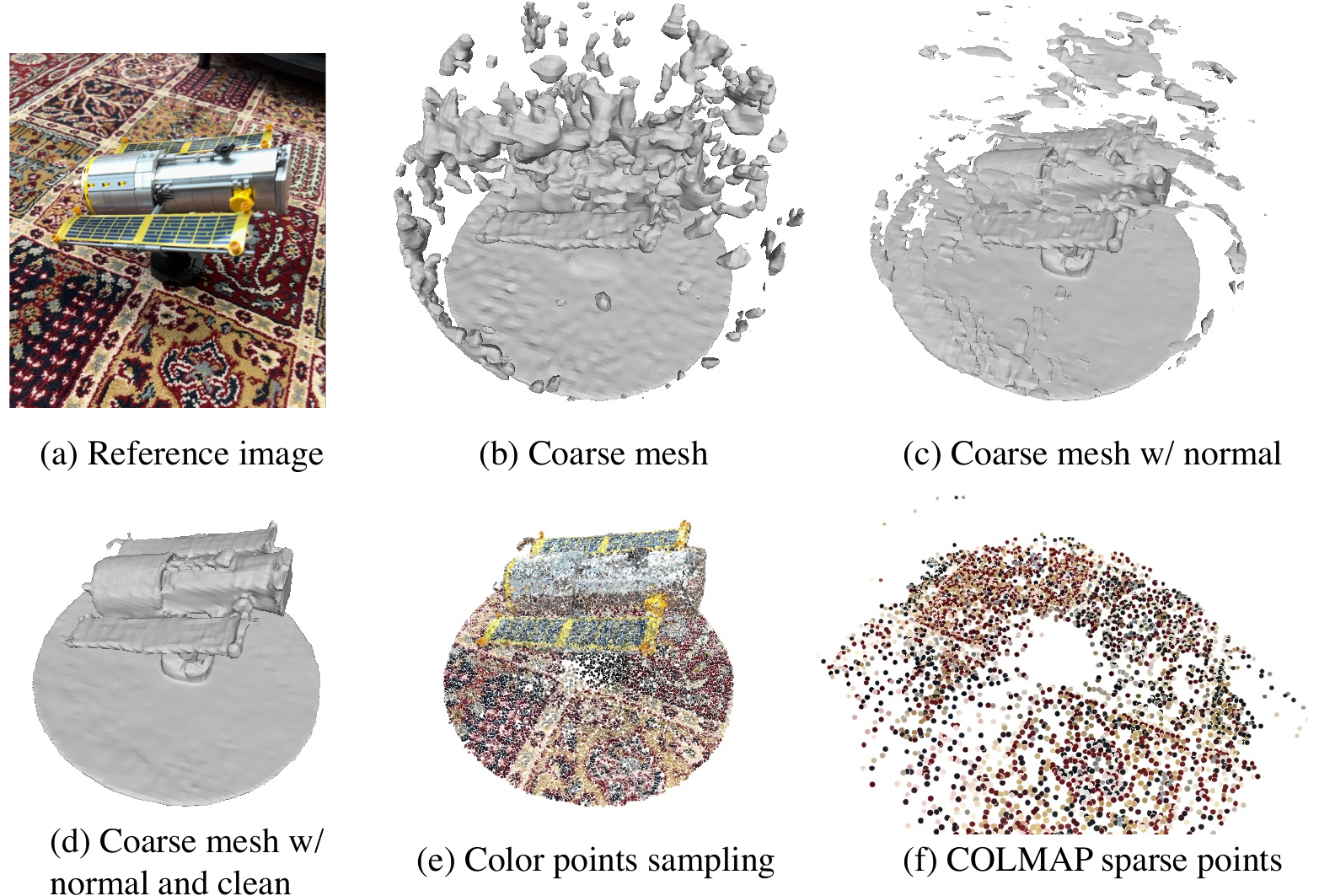}
  \caption{Visualization of our mesh reconstruction, cleaning, and point sampling. (b) Naïve coarse mesh reconstruction following Voxurf~\cite{wu2022voxurf}. (c) Coarse mesh reconstructed with our proposed normal loss, reducing floaters. (d) Post-processed mesh with both normal loss and our cleaning methods. (e) Our sampled point clouds used for initializing 3DGS. (f) COLMAP-estimated point clouds, typically used for 3DGS initialization.}
  \label{fig:method_part1}
\end{figure}

3DGS~\cite{kerbl20233d} typically initializes with sparse point clouds estimated by COLMAP~\cite{snavely2006sfm}, which are often inaccurate or missing in low-texture or little over-lapping regions.
% —see \figref{fig:method_part1}~(f). 
% Limited viewpoint images further hinder repositioning, leading to inaccurate global geometry reconstruction and poor novel view rendering.
To address this, we propose initializing 3DGS by uniformly sampling points from a mesh surface derived from a SDF representation, ensuring high-quality novel view rendering while preserving global geometry.
Below, we detail our proposed method for mesh reconstruction, mesh cleaning, and point cloud sampling. A visual example of the reconstructed meshes and sampled points is shown in \figref{fig:method_part1}.

\paragraph{Coarse mesh reconstruction.} 
Given \( M \) sparse images \( \{\mathcal{I}\} \) and their camera poses \( \{\boldsymbol{\pi}\} \), our objective is to reconstruct a 3D surface for sampling points.
% For our setup, we use sparse input, setting \( M = 10 \) views for 360-degree scenes (e.g., the MobileBrick dataset) and \( M = 5 \) views for narrow-view scenes (e.g., DTU). 
As our focus is on robust global geometry rather than highly accurate surfaces, and to ensure efficient mesh reconstruction, we adopt the coarse-stage surface reconstruction from Voxurf~\cite{wu2022voxurf}.
Specifically, we use a grid-based SDF representation $\boldsymbol{V}^{(\mathrm{sdf})}$ for efficient mesh reconstruction.
For each sampled 3D point \( \mathbf{x} \in \mathbb{R}^3 \), the grid outputs the corresponding SDF value: 
% \begin{equation}
$\boldsymbol{V}^{(\mathrm{sdf})}: \mathbb{R}^3 \rightarrow \mathbb{R}$.
% \end{equation}
We use differentiable volume rendering to render image pixels $\hat{C}(r)$ and employs image reconstruction loss to supervise.
% the two-stage training procedure.
The loss function $\mathcal{L}$ is formulated as:
\begin{equation}
\mathcal{L}=\mathcal{L}_{\text {recon}}+\mathcal{L}_{TV}\left(V^{(\mathrm{sdf})}\right)+ \mathcal{L}_{\text {smooth }}\left(\nabla V^{(\mathrm{sdf})}\right),
\label{eqn:coarse}
\end{equation}

where the reconstruction loss $\mathcal{L}_{\text {recon}}$ calculates photometric image rendering loss, originating from both the $g_{geo}$ and $g_{feat}$ branches. The $\mathcal{L}_{TV}$ encourages a continuous and compact geometry, while the smoothness regularization $\mathcal{L}_{\text {smooth}}$ promotes local smoothness of the geometric surface.
We refer to Voxurf~\cite{wu2022voxurf} for the detailed implementation of the loss functions.
% As our focus is on robust global geometry rather than highly accurate surfaces, and to ensure efficient mesh reconstruction, we adopt the coarse-stage reconstruction setting from Voxurf~\cite{wu2022voxurf}, 
The coarse reconstruction typically completes in 15 minutes in our experiments.

% A visual comparison of generated meshes is shown in \figref{fig:method_part1}~(c) and (d).

Due to the limited number of training views, the learned grid often exhibits floating artifacts, as shown in \figref{fig:method_part1}~(b), which leads to incorrect point sampling. To mitigate this, we introduce a normal consistency loss to improve training stability, effectively reducing floaters and smoothing the geometric surface.
Our approach leverages the predicted monocular surface normal \( \hat{N}(\textbf{r}) \) from the Metric3D model~\cite{yin2023metric3d} to supervise the volume-rendered normal \( \bar{N}(\textbf{r})\) in the same coordinate system. The formulation is:
% 插入具体的voxurf 公式以及 loss
% \begin{equation}
% {N}(\textbf{r})_{\text{gt}} = \left(\boldsymbol{\pi_k}{[:3, :3]}\right) \cdot {N}(\textbf{r}),
% \end{equation}
% where:
% \begin{equation}
% % \mathbf{R} = \left(c2w_{[:3, :3]}\right)^T
% \mathbf{R} = 
% \end{equation}
\begin{equation}
% L_{\text{normal}} = \sum_{\mathbf{r} \in \mathcal{R}} (\left\lVert \hat{{N}}(\mathbf{r}) - N(\textbf{r})_\text{pred} \right\rVert_1)
\mathcal{L}_{\text{normal}} = \sum \big( \Vert \hat{{N}}(\mathbf{r}) - {\bar{N}}(\mathbf{r}) \Vert_1 \big).
% _{\mathbf{r} \in \mathcal{R}}
\end{equation}
%  + \left\lVert 1 - \hat{{N}}(\mathbf{r})^\top N(\textbf{r})_\text{p} \right\rVert_1
We integrate this loss with \equref{eqn:coarse} during training to effectively remove floaters. \figref{fig:method_part1}~(c) shows a coarse mesh reconstructed with the normal loss, demonstrating improved surface smoothness and reduced artifacts.

\paragraph{Mesh cleaning.}
Even though the proposed normal loss significantly reduces floaters, some still persist, adding noise to the subsequent 3DGS initialization. To mitigate this, we apply a mesh cleaning step that refines the coarse mesh by removing non-main components.
Specifically, we first use Marching Cube algorithm~\cite{we1987marching} to extract triangle mesh
$\mathcal{M} = (\mathcal{V}, \mathcal{F})
$ from SDF grid $\boldsymbol{V}^{(\mathrm{sdf})}$. Then
we cluster the connected mesh triangles to \( \{\mathcal{F}_i\} \), identify the largest cluster index:
$|\mathcal{F}_{i_{\text{max}}}| = \max(|\mathcal{F}_i|)$
and get remove parts
\begin{equation}
\mathcal{F}_{\text{remove}} = \{f\in \mathcal{F}\mid f\notin \mathcal{F}_{i_{\text{max}}} \}.
\end{equation}
Finally, we filter the floaters \( \mathcal{F}_{\text{remove}} \) from \( \mathcal{M} \), resulting in $\mathcal{M}_1 = \mathcal{M} \setminus \mathcal{F}_{\text{remove}}$.
\figref{fig:method_part1}~(d) illustrates the refined mesh after applying our cleaning method.

% \begin{figure}
%   \centering
%   \includegraphics[width = 0.5\textwidth]{figure/mesh_algorithm_final2.pdf}
%   \caption{Coarse mesh post-processing pipeline.}
%   \label{fig:mesh_processing}
% \end{figure}

\paragraph{Sampling surface points for 3DGS.}
% 模块结构/流程，motivation，优势
% \figref{fig:mesh_processing} illustrates the post-processing pipeline for coarse meshes and uniform sampling.

% Since the mesh obtained from Marching Cubes includes regions invisible from the training views, directly sampling points from the mesh surface would introduce noise into 3DGS. 
% To address this, we propose a depth-based sampling strategy.
% First, we project the reconstructed mesh onto the training views using their known camera poses and obtain depth maps $\left\{\mathcal{D}\right\}$.
% As the depth map is projected from a 3D mesh,
% They are naturally consistent over multiple views.
% Then, we randomly sample points from valid depth regions. These pixels ${(u,v)}$, along with depth values ${d(u,v)}$, are projected to colorized 3D points $\mathbf{P}=\{ (x_i, y_i, z_i) \mid i = 1, 2, \ldots, N \}$, formulated as:  
% % $p_w = \left[
% % x_w y_w z_w
% % \right]^{\top}=\boldsymbol{\pi_i} \mathbf{K}^{-1}\left[
% % u v 1\right]^{\top}$.
% \begin{equation}
% \begin{bmatrix} x_i & y_i & z_i \end{bmatrix} = \boldsymbol{\pi_k} \mathbf{K}^{-1} \begin{bmatrix} d \cdot u & d \cdot v & d \end{bmatrix}^T.
% \end{equation}
% This ensures that the sampled points are uniformly distributed on the object’s surface while remaining visible in the training views, leading to a more stable and accurate 3DGS initialization.

Since the mesh obtained from Marching Cubes includes regions that are invisible from the training views, directly sampling points from the mesh surface can introduce noise into 3DGS. To mitigate this, we propose a depth-based sampling strategy.
First, we project the reconstructed mesh onto the training views using their known camera poses to generate depth maps \( \{\mathcal{D}\} \). Since these depth maps originate from a 3D mesh, they maintain multi-view consistency. We then randomly sample points from valid depth regions, ensuring they correspond to visible object surfaces. The sampled pixels \( (u,v) \), along with their depth values \( d(u,v) \), are back-projected to colorized 3D points \( \mathbf{P} = \{ (x_i, y_i, z_i) \mid i = 1, 2, \dots, N \} \) using the following formulation:
\begin{equation}
\begin{bmatrix} x_i & y_i & z_i \end{bmatrix} = \boldsymbol{\pi_k} \mathbf{K}^{-1} \begin{bmatrix} d \cdot u & d \cdot v & d \end{bmatrix}^T.
\end{equation}
This approach ensures that the sampled points are uniformly distributed on the object’s surface while remaining visible in the training views, leading to a more stable and accurate 3DGS initialization.
As our reconstructed mesh primarily covers foreground regions, we combine our sampled point cloud with COLMAP sparse points when rendering background regions, serving as the initialization for 3DGS.
\figref{fig:method_part1}~(e) and (f) illustrate our sampled point clouds and COLMAP-estimated point clouds, respectively.

% Finally, we combine it (\figref{fig:method_part1}~(e)) with sparse COLMAP points (\figref{fig:method_part1}~(f)), serving as the initialization for 3D Gaussian optimization.\\
% 采样的公式/算法图/流程图

\subsection{3DGS for Enhanced SDF}\label{method:rendering}

We argue that the primary bottleneck for SDF-based mesh reconstruction is insufficient supervision due to limited training views. To address this, we generate additional novel viewpoint images using a 3DGS-based method and combine them with the original sparse views to enhance the training of SDF-based reconstruction.

\paragraph{Rendering novel viewpoint images.}
% 只训练7k次
% \begin{figure}
%   \centering
%   \includegraphics[width = 0.5\textwidth]{figure/cam_pos.pdf}
%   \caption{cam position}
%   \label{fig:cam_position}
% \end{figure}
\begin{figure}
  \centering
  \includegraphics[width = 0.49\textwidth]{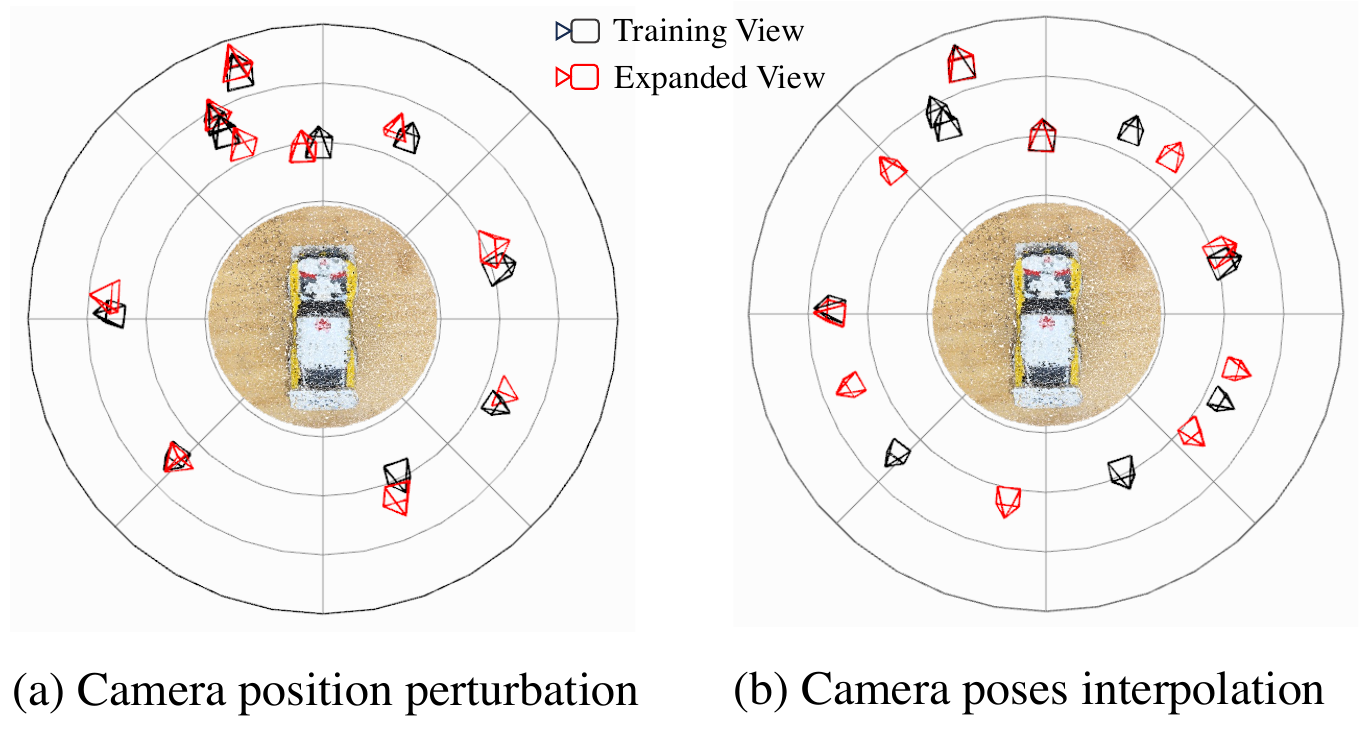}
  \caption{Top-view visualization of pose expansion strategies.}
  \label{fig:cam_position}
\end{figure}
We utilize the improved 3DGS, initialized with our proposed mesh-based point sampling method, to render images. Thanks to our robust and dense point initialization, the 3D Gaussian \( \mathcal{G} \) can converge after \( 7k \) iterations in just 5 minutes, yielding
% \begin{equation}
$\mathcal{G} = f(\mathbf{P}, \{I\}, \{\boldsymbol{\pi}\})$.
% \end{equation}
Given new camera poses \( \{\boldsymbol{\pi_{\text{new}}}\} \), the 3D Gaussian \( \mathcal{G} \) can be projected to generate novel-view images as follows:
\begin{equation}
\{\mathcal{I}_{\text{new}}\} = \text{Splat}(\mathcal{G}, \{\boldsymbol{\pi_{\text{new}}}\}).
\end{equation}
The newly rendered images \( \{\mathcal{I}_{\text{new}}\} \) are combined with the input images \( \{\mathcal{I}\} \)  to train the SDF-based mesh reconstruction.
The key challenge lies in selecting new camera viewpoints $\{\boldsymbol{\pi_{\text{new}}}\}$ that best enhance surface reconstruction:
\begin{equation}
    \{\boldsymbol{\pi_{\text{new}}}\}=g\left(\left\{\boldsymbol{\pi}\right\}\right)
\end{equation}
where $g$ is our pose expansion strategy. 
% To prevent new viewpoints from deviating excessively from the original camera pose distribution, which could cause blurred images or reduce the foreground subject's prominence, we explored two methods for generating new camera poses.
To ensure new viewpoints remain consistent with the original pose distribution and avoid excessive deviation that could blur or diminish the foreground, we explore two methods for generating new camera poses.
\figref{fig:cam_position} shows the generated pose position.

\paragraph{Camera position perturbation.} 
% A delta offset is applied to the original camera positions, and new camera extrinsics are calculated based on the target viewpoint. This method allows control over new viewpoints around the original distribution, ensuring that the new views do not deviate too far, thus maintaining high-quality rendering for downstream reconstruction.
To generate new camera positions while preserving proximity to the original distribution, a perturbation $\Delta \mathbf{p}$ is applied to the initial camera positions $\left\{\boldsymbol{c}\right\}$. The new camera centers $\left\{\boldsymbol{c}^{\prime}_m\right\}$ are computed:
\begin{equation}
\boldsymbol{c}^{\prime}_m =\boldsymbol{c}+\Delta \mathbf{p},
\end{equation}
where $\Delta \mathbf{p}=(\Delta x, \Delta y, \Delta z)$ represents a controlled offset vector designed to modulate the new viewpoints. 
% This method keeps the extended positions close to the original configuration, avoiding deviations that could reduce image quality. 
% By maintaining a balanced perturbation, high-fidelity rendering is preserved, supporting accurate downstream mesh refinement.
\paragraph{Camera pose interpolation.} 
% This method generates smooth, continuous interpolated camera poses between specified viewpoints. 
Our method takes a set of camera rotation matrices $\left\{\mathbf{R}\right\}$ and camera positions $\left\{\boldsymbol{c}\right\}$ as input.
To generate smooth transitions between viewpoints, we employ cubic spline interpolation~\cite{de1978splines}.
This approach interpolates both camera positions and orientations, producing interpolated camera centers $\left\{\boldsymbol{c}^{\prime}_m\right\}$ and rotation matrices 
$\left\{\mathbf{R}^{\prime}_m\right\}$ that ensure visual continuity and positional coherence.
By maintaining these properties, the newly generated camera poses facilitate high-quality transitions, making them well-suited for 3D mesh reconstruction.
The visualizations of the images generated from new viewpoints can be found in Fig. 2 of the supplementary material.  
% By interpolating both camera positions and orientations, the function produces interpolated camera centers $\left\{\boldsymbol{c}^{\prime}_m\right\}$ and rotation matrices $\left\{\mathbf{R}^{\prime}_m\right\}$  that maintain visual continuity and positional coherence and ensures that the new camera poses facilitate high-quality transitions suitable for 3D mesh reconstruction.

\paragraph{Refining surface reconstruction.}
We reuse the reconstructed coarse mesh and refine it with the original and expanded novel viewpoint images. 
Following the fine-stage reconstruction of Voxurf~\cite{wu2022voxurf}, we increase the grid resolution and introduce a dual color network and hierarchical geometry features for detailed surface reconstruction.

% \subsection{Iterative progressive loop}
\subsection{Cyclic Optimization}

% The two key components introduced above form a complete pipeline for sparse novel view synthesis and reconstruction. In fact, this entire process can be looped, progressively enhancing the 3D representation. The specific steps are as follows:

% 1. Volume-based coarse mesh reconstruction provides the initial point cloud.
% 2. Optimize 3DGS to render novel views via frame interpolation.
% 3. Use both the novel views and original images for mesh reconstruction.
% 4. The improved mesh offers a more refined point cloud initialization.
% 5. The higher-quality 3DGS training continues to generate novel views, further aiding subsequent geometric reconstructions. This loop repeats.

\setlength{\tabcolsep}{3.05pt}
\begin{table*}[htbp]
  \centering
  \small
  \caption{Surface reconstruction and novel view synthesis results on MobileBrick. The results are averaged over all 18 test scenes with an initial input of 10 images per scene. PSNR-F is computed only on foreground regions. The best results are \textbf{bolded}.}
    \begin{tabular}{l|c|c|c|r|r|r|c|c|c|c}
    %\toprule
    \toprule
          & \multicolumn{7}{c|}{Mesh Reconstruction}              & \multicolumn{2}{c|}{Rendering} & \multicolumn{1}{c}{\multirow{3}[6]{*}{Time}}\\
    % \midrule
    \cmidrule{1-10}
          & \multicolumn{3}{c|}{$\sigma=2.5mm$} & \multicolumn{3}{c|}{$\sigma=5mm$} & \multirow{2}[4]{*}{CD (mm)↓} & 
          \multirow{2}[4]{*}{PSNR↑} & \multirow{2}[4]{*}{PSNR-F↑} \\
\cmidrule{1-7}          & Accu.(\%)↑ & Recall(\%)↑ & F1↑   & \multicolumn{1}{c|}{Accu.(\%)↑} & \multicolumn{1}{c|}{Recall(\%)↑} & \multicolumn{1}{c|}{F1↑} &       &       &  \\
    \midrule
    \midrule
    Voxurf~\cite{wu2022voxurf} & 62.89 & 62.54 & 62.42 & \multicolumn{1}{c|}{80.93} & \multicolumn{1}{c|}{80.61} & \multicolumn{1}{c|}{80.38} & 13.3  & 14.34 & 18.34 & 55 mins\\
    \midrule
    MonoSDF~\cite{Yu2022MonoSDF} & 41.56 & 32.47 & 36.22 & \multicolumn{1}{c|}{57.88} & \multicolumn{1}{c|}{48.19} & \multicolumn{1}{c|}{52.21} & 37.7  & 14.71  & 15.42 & 6 hrs\\
    \midrule
    
    2DGS~\cite{Huang2DGS2024}  & 49.83 & 45.32 & 47.10 & \multicolumn{1}{c|}{72.65} & \multicolumn{1}{c|}{64.88} & \multicolumn{1}{c|}{67.96}   & 14.8 & 17.12 & 18.52 & 10 mins \\
    \midrule
    GOF~\cite{Yu2024GOF}   & 50.24 & 61.11 & 54.96 & \multicolumn{1}{c|}{74.99} & \multicolumn{1}{c|}{82.68} & \multicolumn{1}{c|}{78.16} & 11.0 & 16.52 & 18.36 & 50 mins \\
    \midrule
    3DGS~\cite{kerbl20233d}  & \textbackslash{} & \textbackslash{} & \textbackslash{} & \multicolumn{1}{c|}{\textbackslash{}} & \multicolumn{1}{c|}{\textbackslash{}} & \multicolumn{1}{c|}{\textbackslash{}} & \textbackslash{} & 17.19 & 19.12 & 10 mins \\
    \midrule
    SparseGS~\cite{xiong2023sparsegs} & \textbackslash{} & \textbackslash{} & \textbackslash{} & \multicolumn{1}{c|}{\textbackslash{}} & \multicolumn{1}{c|}{\textbackslash{}} & \multicolumn{1}{c|}{\textbackslash{}} & \textbackslash{} & 16.93 & 18.74 & 30 mins\\
    \midrule
    Ours  & 68.36 & \textbf{69.79} & 68.97 & \multicolumn{1}{c|}{86.79} & \multicolumn{1}{c|}{\textbf{86.82}} & \multicolumn{1}{c|}{86.65} & \textbf{9.7}   & 17.48 & 20.45 & 1 hr \\
\midrule
    Ours (Two cycles)  & \textbf{69.61} & 68.89 & \textbf{69.14} & \multicolumn{1}{c|}{\textbf{87.79}} & \multicolumn{1}{c|}{85.93} & \multicolumn{1}{c|}{\textbf{86.74}} & 9.9   & \textbf{17.58} & \textbf{20.55} & 1.6 hr \\
    \bottomrule
 %   \bottomrule
    
    \end{tabular}%
  \label{tab:mobilebrick_results}%
\end{table*}%

\begin{table*}[htbp]
  \centering
  \setlength{\tabcolsep}{3pt} 
  % \footnotesize
  \small
  \caption{Surface reconstruction results on DTU with 5 input views. Values indicate Chamfer Distance in millimeters (mm). "-" denotes failure cases where COLMAP could not generate point clouds for 3DGS initialization. GSDF-10 is reported with 10 input images, as it fails in sparser settings.
The best results are \textbf{bolded}, while the second-best are \underline{underlined}. } 
    \begin{tabular}{lcccccccccccccccccc}
   % \toprule
    \toprule
    Scan & 24    & 37    & 40    & 55    & 63    & 65    & 69    & 83    & 97    & 105   & 106   & 110   & 114   & 118   & 122   & Mean & Time \\
    \midrule
    Voxurf~\cite{wu2022voxurf} & 2.74  & 4.50  & 3.39  & 1.52  & 2.24  & 2.00  & 2.94  & \textbf{1.29}  & 2.49  & 1.28  & 2.45  & 4.69  & 0.93  & 2.74  & \underline{1.29}  & 2.43  & 50 mins \\
    MonoSDF~\cite{Yu2022MonoSDF} & \textbf{1.30}   & \underline{3.45}     & \textbf{1.45}   & \textbf{0.61}  & \textbf{1.43}  & \textbf{1.17}  & \textbf{1.07}  & \underline{1.42}  & \textbf{1.49}  & \textbf{0.79}  & 3.06  & \underline{2.60}  & \underline{0.60}  & 2.21  & 2.87  & 
    \underline{1.70} & 6 hrs  \\
    \midrule

SparseNeuS~\cite{long2022sparseneus} & 3.57  & 3.73  & 3.11  & 1.50 & 2.36  &  2.89 & 1.91 & 2.10   & 2.89  &  2.01 &  \underline{2.08} &  3.44  & 1.21 &  2.19 &  2.11 &  2.43 & Pretrain + 2 hrs ft \\
    \midrule

    2DGS~\cite{Huang2DGS2024}  & 4.26  & 4.80  & 5.53  & 1.50  & 3.01  & 1.99  & 2.66  & 3.65  & 3.06  & 2.54  & 2.15  & -   & 0.96  & \underline{2.17}  & 1.31  & 2.84  & 6 mins \\
    GOF (TSDF)~\cite{Yu2024GOF} & 7.30  & 5.80  & 6.03  & 2.79  & 4.23  & 3.41  & 3.44  & 4.37  & 3.75  & 2.99  & 3.19  & -   & 2.64  & 3.67  & 2.25  & 4.03  & 50 mins \\
    GOF~\cite{Yu2024GOF} & 4.37  & 3.68  & 3.84  & 2.29  & 4.40  & 3.28  & 2.84  & 4.64  & 3.40  & 3.76  & 3.56  & -   & 3.06  & 2.95  & 2.91  & 3.55 & 50 mins \\
    \midrule
GSDF-10~\cite{yu2024gsdf} &   6.89  &  6.82     & 7.97  & 6.54  & 5.22  & 1.91  & 5.56  &  4.38   & 7.01  & 3.69  & 6.33  & 6.33  & 3.95  & 6.30   & 2.09  &  5.40  & 3 hrs \\

    Ours  & \underline{1.55}  & \textbf{2.64}  & \underline{1.52}  & \underline{1.40}  & \underline{1.51}  & \underline{1.46}  & \underline{1.23}  & 1.43  & \underline{1.82}  & \underline{1.19}  & \textbf{1.49}  & \textbf{1.80}  & \textbf{0.54}  & \textbf{1.19}  & \textbf{1.04}  & \textbf{1.45} & 1 hr \\
    \bottomrule
 %   \bottomrule
    \end{tabular}%
  \label{tab:dtu_results}%
\end{table*}%

% This iterative process effectively combines the strengths of both methods, allowing them to complement each other through repeated cycles, thus improving both the quality of the reconstruction and novel view synthesis.
% When one iteration of coarse mesh-rendering-fine mesh refinement is completed, we regard it as a single optimization cycle. By continuing this iterative process, which stabilizes into a mesh-rendering-mesh-rendering sequence, the approach effectively harnesses the strengths of both methods. This cyclic refinement allows them to complement each other through repeated iterations, thereby enhancing the quality of the reconstruction and improving novel view synthesis.

We propose an interactive optimization process, which begins by generating an initial coarse mesh $\mathcal{M}^{(0)}$.
Then, in each iteration $n$, the process follows two steps:\\
\indent 1. Rendering Step: We optimize a 3DGS model for rendering novel view images, which is initialized by sampling points from the current coarse mesh $\mathcal{M}_c^{(n)}$, represented by:
\begin{equation}
\mathcal{I}^{(n)}=\mathcal{R}\left(\mathcal{M}_c^{(n)}\right)
\end{equation}
\indent 2. Meshing Step: We refine the current mesh by fine-tuning it using both the newly rendered images and the original input images:
\begin{equation}
\mathcal{M}_f^{(n)}=\mathcal{O}\left(\mathcal{M}_c^{(n)}, \mathcal{I}^{(n)}\right)
\end{equation}
where 
$\mathcal{O}$ represents the SDF grid optimization.
Then, we update the refined mesh:
\begin{equation}
\mathcal{M}_c^{(n+1)}=\mathcal{M}_f^{(n)}.
\end{equation}
By iterating this process, our method allows SDF-based reconstruction and 3DGS-based rendering to complement each other, improving both reconstruction accuracy and novel view synthesis. To balance efficiency and accuracy, we typically perform only one iteration.

\section{Experiments}
\label{sec:experiment}

\subsection{Experimental Setup}
\paragraph{Datasets.}
We conduct a comprehensive evaluation of the proposed method on the MobileBrick
\cite{li2023mobilebrick} and DTU \cite{DTU2014} datasets. MobileBrick is a multi-view RGB-D dataset captured on a mobile device, providing precise 3D annotations for detailed 3D object reconstruction. Unlike the DTU dataset, which is captured in a controlled lab environment, MobileBrick represents more challenging, real-world conditions, making it more reflective of everyday scenarios.
Following previous methods~\cite{wu2022voxurf, wang2021neus, li2023mobilebrick}, we use 15 test scenes from DTU and 18 test scenes from MobileBrick for evaluation. In the MobileBrick dataset, each scene consists of 360-degree multi-view images, from which we sample 10 images with 10\% overlap for sparse view reconstruction. In contrast, the DTU dataset, with higher overlap, is sampled with 5 frames per scene. We also present reconstruction results for the little-overlapping 3-view setting in the supplementary materials. For fair comparison, 3DGS-based methods are initialized using point clouds from COLMAP\cite{schoenberger2016sfm} with ground-truth poses. The selected images and poses are used for 3D reconstruction, while the remaining images serve as a test set for evaluating novel view rendering.

\paragraph{Baselines.}
We compare our proposed method with both SDF-based and 3DGS-based approaches for surface reconstruction. The SDF-based methods include MonoSDF\cite{Yu2022MonoSDF}, Voxurf\cite{wu2022voxurf}, and SparseNeuS~\cite{long2022sparseneus}, which is pre-trained on large-scale data.
The 3DGS-based methods include 2DGS\cite{Huang2DGS2024} and GOF\cite{Yu2024GOF}.
Additionally, we compare with GSDF~\cite{yu2024gsdf}, which integrates both SDF and 3DGS, similar to our approach, but is designed for dense-view settings.
For novel view rendering, we evaluate all these methods along with 3DGS\cite{kerbl20233d} and SparseGS\cite{xiong2023sparsegs}.

\paragraph{Evaluation metrics.}
We follow the official evaluation metrics on MobileBrick, reporting Chamfer Distance, precision, recall, and F1 score at two thresholds: $2.5mm$ and $5mm$.
For the DTU dataset, we use Chamfer Distance as the primary metric for surface reconstruction.
To evaluate novel view rendering performance, we report PSNR for full images and PSNR-F, which is computed only over foreground regions. In each scene, we train models using sparse input images and test on all remaining views. The final result is averaged over all evaluation images.

\paragraph{Implementation details.}
We set the voxel grid resolution to \(96^3\) during coarse mesh training, requiring approximately 15 minutes for 10k iterations. The weight of the proposed normal loss is set to 0.05, while all other parameters follow Voxurf~\cite{wu2022voxurf}.
Next, we train 3DGS~\cite{kerbl20233d} for 7k iterations, which takes around 5 minutes, and render 10 new viewpoint images within 30 seconds.
After expanding the training images, we increase the voxel grid resolution to \(256^3\) and train for 20k iterations, taking approximately 40 minutes. Thus, a complete optimization cycle takes roughly 1 hour.

\subsection{Comparisons}

\begin{figure*}
  \centering
\includegraphics[width = 0.96\textwidth]{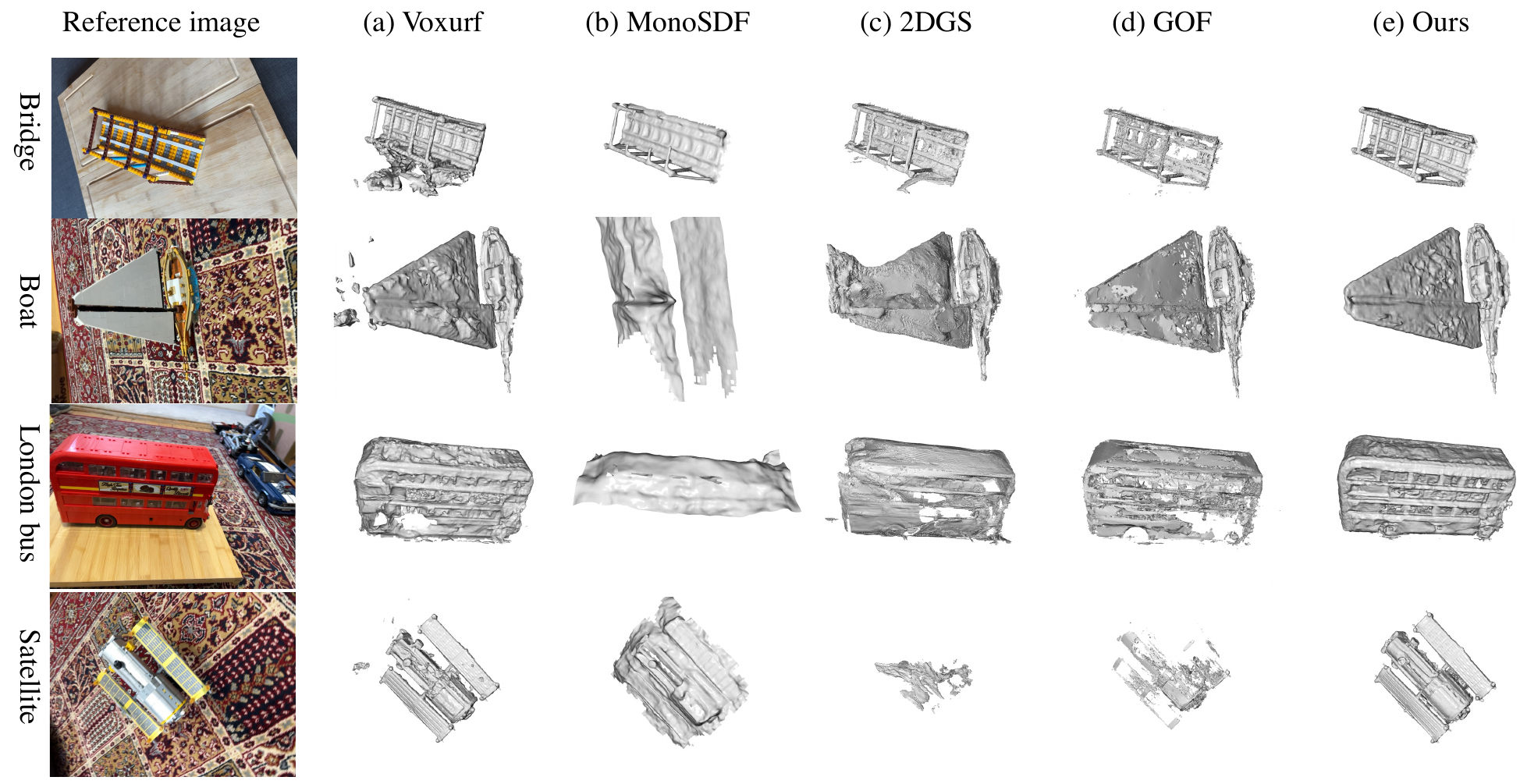}
  \caption{Qualitative mesh reconstruction comparisons on MobileBrick. See more visual results in supplementary material.}
  \label{fig:mesh_vis}
\end{figure*}

\begin{figure*}
  \centering
\includegraphics[width = 0.97\textwidth]{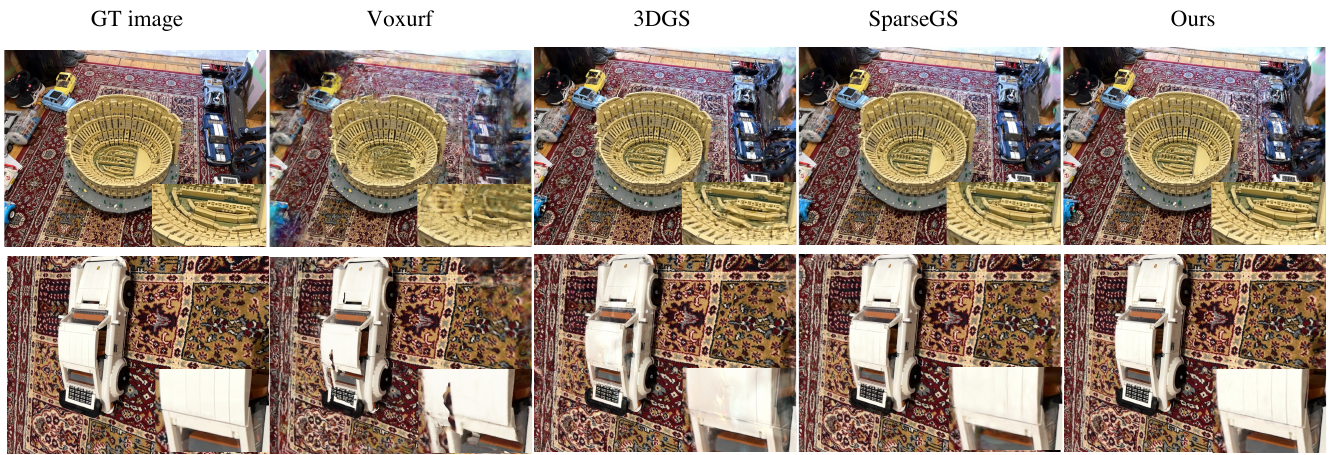}
  \caption{Qualitative novel view synthesis comparisons on MobileBrick.}
  \label{fig:nvs_vis}
\end{figure*}

\paragraph{Results on MobileBrick.}
\tabref{tab:mobilebrick_results} presents a quantitative comparison of our method against previous approaches. The results show that Voxurf~\cite{wu2022voxurf}, which utilizes an SDF-based representation, outperforms 2DGS~\cite{Huang2DGS2024} and GOF~\cite{Yu2024GOF} (both 3DGS-based methods) in surface reconstruction metrics, particularly in terms of the F1 score. However, all 3DGS-based methods achieve notably better novel view rendering performance, as evidenced by their higher PSNR values compared to Voxurf.
A visual comparison is illustrated in \figref{fig:mesh_vis} and \figref{fig:nvs_vis}.
BBy leveraging the strengths of both SDF and 3DGS representations, our method achieves state-of-the-art performance in surface reconstruction and novel view synthesis. To balance efficiency and performance, we adopt a single-cycle approach in practice.
\paragraph{Results on DTU.}
\tabref{tab:dtu_results} presents surface reconstruction results on the DTU dataset, which is particularly challenging due to the use of only 5 uniformly sampled frames for reconstruction. SparseNeuS~\cite{long2022sparseneus} is a pre-trained model that requires an additional 2 hours of fine-tuning.
COLMAP fails to generate sparse point clouds for scene 110, preventing 3DGS initialization. GSDF~\cite{yu2024gsdf} struggles in sparse-view settings, so we train it on 10 images. Despite these challenges, our method achieves robust reconstruction and significantly outperforms other approaches.

% \textbf{Mesh Reconstruction Results.}
% On MobileBrick dataset, we evaluated the mesh reconstruction quality from 10 sparsely distributed 360-degree views on $18$ test scenes. The mean quantitative results of $18$ scenes are shown in Table 1. Compared to other advanced 3D mesh reconstruction techniques, our method demonstrates a clear advantage, achieving the highest precision, recall, and F1 score under both thresholds, and the lowest error in Chamfer Distance for $9.7mm$.\\
% \textbf{Rendering Results.}
% We also evaluated the novel view synthesis performance of various methods on the $18$ test scenes of MobileBrick dataset. Using the same 10 sparsely distributed 360-degree views for training, we tested on the remaining images from each scene. The evaluation metrics include the full-image PSNR and the PSNR for the segmented foreground, as we are particularly focused on the reconstruction quality of the foreground objects. Similarly, our method achieved the highest scores for both metrics. Notably, for MASK PSNR, our results exceeded other methods by at least one point.

\subsection{Ablations}

% \begin{figure}
%   \centering
%   \includegraphics[width = 0.5\textwidth]{figure/Ablation_1.pdf}
%   \caption{Improvement of chamfer distance given different number of input views.}
%   \label{fig:ablation_input}
% \end{figure}

\begin{table}[tbp]
  \centering
  \small
  \caption{Surface reconstruction results with varying numbers of input views on MobileBrick (porsche) and DTU (scan69). The Baseline represents a pure SDF-based reconstruction without the assistance from 3DGS. $\delta$ indicates the improvement.}
    \begin{tabular}{c|ccc|ccc}
   % \toprule
    \toprule
          & \multicolumn{3}{c|}{MobileBrick / F1 score} & \multicolumn{3}{c}{DTU / CD} \\
    Input& Baseline & Ours  & $\delta$ & Baseline & Ours  & $\delta$ \\
    \midrule
    5     & 33.50  & \textbf{43.11} & +9.61  & 2.940 & \textbf{1.230} & -1.710 \\
    10    & 59.66 & \textbf{62.37} & +2.71  & 1.362 & \textbf{1.165} & -0.197 \\
    20    & 63.18 & \textbf{63.88} & +0.7   & 1.043 & \textbf{0.965} & -0.078 \\
    \bottomrule
  %  \bottomrule
    \end{tabular}%
  \label{tab:ablation_input_number}%
\end{table}%

% \begin{table}[tbp]
%   \centering
%   \footnotesize
%   \caption{Ablations studies on effectiveness of our proposed modules on MobileBrick test scenes.}
%     \begin{tabular}{l|c|c|c|c}
%     \toprule
%     \toprule
%           & \multicolumn{2}{c|}{Meshing} & \multicolumn{2}{c}{Rendering} \\
% \cmidrule{1-5}  &  F1↑ & CD↓  & PSNR↑ & PSNR MASK↑  \\
%     \midrule
%     Mesh w/o 3DGS & 62.43 & 13.3 & - & - \\
%     3DGS w/o Mesh & -  & - & 17.19 & 19.12 \\
%     Ours (one cycle) & 68.97 & \textbf{9.7} & 17.48 & 20.45 \\
%     Ours (two cycles) & \textbf{69.14} & 9.9 & \textbf{17.58} & \textbf{20.55} \\
%     \bottomrule
%     \bottomrule
%     \end{tabular}%
%   \label{tab:connectivity}%
% \end{table}%

% \paragraph{Number of input images.}
% We evaluate our method across different levels of sparsity, using up to 20 images per scene (additional results are provided in the supplementary materials). \tabref{tab:ablation_input_number} summarizes the results on both MobileBrick and DTU, demonstrating consistent improvements across all settings and highlighting the effectiveness of our approach, especially in sparser scenarios.

\paragraph{Efficacy of 3DGS for Improving SDF.}
\tabref{tab:ablation_input_number} compares our method with a pure SDF-based reconstruction baseline at different sparsity levels, using up to 20 images per scene. The results on MobileBrick and DTU validate the effectiveness of our 3DGS-assisted SDF approach.
% We compare our method with a pure SDF-based reconstruction baseline to validate the effectiveness of our 3DGS-assisted SDF approach. Additionally, we evaluate our method across different levels of sparsity, using up to 20 images per scene. \tabref{tab:ablation_input_number} presents results on MobileBrick and DTU, demonstrating consistent improvements across all settings and highlighting the effectiveness of our approach.
More results are provided in the supplementary material.

\begin{table}[tb]
  \centering
  \small
  \caption{3DGS rendering results with different initializations, averaged across all 18 MobileBrick test scenes.}
    \begin{tabular}{c|c}
 %   \toprule
    \toprule
    Method & Foreground PSNR \\
    \midrule
    3DGS (COLMAP) & 19.13 \\
    % \midrule
    3DGS w/ mesh clean & 19.88 \\
    % \midrule
    3DGS w/ normal and mesh clean & \textbf{20.45} \\
    \bottomrule
 %   \bottomrule
    \end{tabular}%
  \label{tab:mesh_for_3dgs}%
\end{table}%

\begin{figure}[H]
  \centering
  \includegraphics[width = 0.45\textwidth]{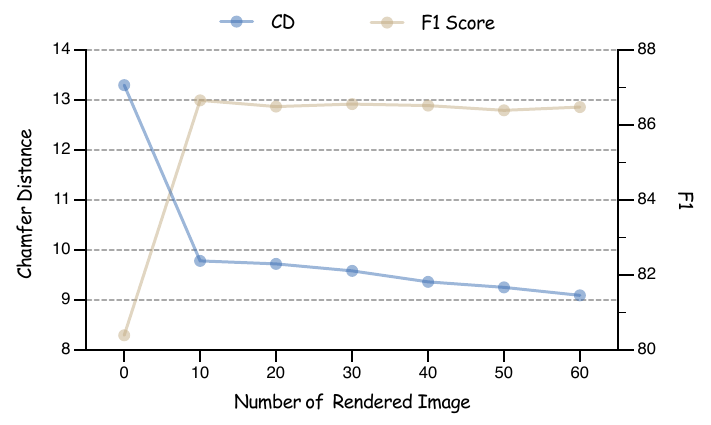}
  \caption{econstruction quality with varying numbers of 3DGS-rendered novel view images from expanded poses, averaged across all 18 MobileBrick test scenes, with an initial input of 10 images.}
  \label{fig:ablation_number}
\end{figure}

\paragraph{Efficacy of SDF for enhancing 3DGS.}

% Our proposed SDF-based method effectively enhances the quality of the 3DGS initialization point cloud.
\tabref{tab:mesh_for_3dgs} compares the novel view rendering results for 3DGS using point clouds initialized with different sampling strategies. The results demonstrate that our proposed mesh cleaning and normal supervision notably improve 3DGS performance.

\paragraph{Number of newly rendered views.}
% \figref{fig:ablation_number} illustrates the impact of the number of newly rendered images on surface reconstruction quality. 
% The results on MobileBrick show that rendering 10 additional novel views yields substantial improvements in Chamfer Distance (CD) and F1, with gains of approximately 26.5\% and 7.2\%, respectively. As the number of novel views increases, the accuracy gains begin to diminish, though additional views continue to slightly improve CD. This suggests that while additional renderings can refine reconstruction, the majority of benefits are achieved with the initial 10 renderings.
\figref{fig:ablation_number} illustrates the impact of the number of newly rendered images on surface reconstruction. On MobileBrick, rendering 10 novel views significantly improves Chamfer Distance (26.5\%) and F1 (7.2\%). As the number of novel views increases, accuracy gains gradually diminish. 
% though additional views continue to provide slight CD improvements. 
This suggests that while additional renderings refine reconstruction, the majority of benefits are achieved with the first 10 rendered images.

% Table generated by Excel2LaTeX from sheet 'Sheet3'
\begin{table}[bt]
  \centering
  \small
  \caption{Ablation study on pose expansion strategies for in MobileBrick (aston) with 10 input images.}.
    \begin{tabular}{c|c|c|c}
 %   \toprule
    \toprule
          & F1↑   & Recall(\%)↑ & CD (mm)↓ \\
    \midrule
    Baseline & 55.8  & 49.9  & 8.7 \\
    \midrule
    % Spherical fitting & 58.6  & 56.7  & 6.6 \\
     Camera position perturbation & 59.9  & 57.4  & 6.6 \\
    Camera poses interpolation & \textbf{60.8}  & \textbf{59.1}  & \textbf{6.4}\\
 %   \bottomrule
    \bottomrule
    \end{tabular}%
  \label{tab:sampling}%
\end{table}%

\paragraph{Different pose expansion strategies.}
% \figref{fig:cam_position} shows a visualization of the proposed strategies. 
\tabref{tab:sampling} summarizes the reconstruction performance with expansion images from different strategies. We double the number of original input camera poses, generating new viewpoints and rendering additional images accordingly.
The two strategies significantly enhance surface reconstruction quality, with camera pose interpolation yielding the greatest improvement.

\section{Conclusion}
\label{sec:Conclusion}
\vspace{-0.8mm}
This paper introduces a novel framework for sparse-view reconstruction, where SDF-based and 3DGS-based representations complement each other to enhance both surface reconstruction and novel view rendering. 
Specifically, our method leverages SDF for modeling global geometry and 3DGS for capturing fine details, achieving significant improvements over state-of-the-art methods on two widely used real-world datasets.

\textbf{Limitation and future work.}
Although our method can theoretically be generalized to any SDF and novel view rendering approaches, our current implementation is built on Voxurf and 3DGS, which were selected for their efficiency-performance trade-off. As a result, our method is currently limited to object-level scenes and struggles with extremely sparse inputs, such as only two images.
In the future, we aim to extend our approach to handle more diverse scenes and further improve its robustness to sparse inputs.

\subsection*{Acknowledgments}

This work was supported by the National Natural Science Foundation of China (No.\ 62206244)
.

{
    \small
    \bibliographystyle{ieeenat_fullname}
    \bibliography{main}

\begin{thebibliography}{59}
\providecommand{\natexlab}[1]{#1}
\providecommand{\url}[1]{\texttt{#1}}
\expandafter\ifx\csname urlstyle\endcsname\relax
  \providecommand{\doi}[1]{doi: #1}\else
  \providecommand{\doi}{doi: \begingroup \urlstyle{rm}\Url}\fi

\bibitem[Bai et~al.()Bai, Huang, Guo, Gong, Li, and Guo]{bai2024360}
Jiayang Bai, Letian Huang, Jie Guo, Wen Gong, Yuanqi Li, and Yanwen Guo.
\newblock 360-gs: Layout-guided panoramic gaussian splatting for indoor roaming.
\newblock In \emph{International Conference on 3D Vision 2025}.

\bibitem[Barron et~al.(2021)Barron, Mildenhall, Tancik, Hedman, Martin-Brualla, and Srinivasan]{barron2021mip}
Jonathan~T Barron, Ben Mildenhall, Matthew Tancik, Peter Hedman, Ricardo Martin-Brualla, and Pratul~P Srinivasan.
\newblock Mip-nerf: A multiscale representation for anti-aliasing neural radiance fields.
\newblock In \emph{Int. Conf. Comput. Vis.}, pages 5855--5864, 2021.

\bibitem[Barron et~al.(2022)Barron, Mildenhall, Verbin, Srinivasan, and Hedman]{barron2022mip}
Jonathan~T Barron, Ben Mildenhall, Dor Verbin, Pratul~P Srinivasan, and Peter Hedman.
\newblock Mip-nerf 360: Unbounded anti-aliased neural radiance fields.
\newblock In \emph{IEEE Conf. Comput. Vis. Pattern Recog.}, pages 5470--5479, 2022.

\bibitem[Bian et~al.(2024)Bian, Bian, Prisacariu, and Torr]{bian2024porf}
Jia-Wang Bian, Wenjing Bian, Victor~Adrian Prisacariu, and Philip Torr.
\newblock Porf: Pose residual field for accurate neural surface reconstruction.
\newblock In \emph{ICLR}, 2024.

\bibitem[Bian et~al.(2023)Bian, Wang, Li, Bian, and Prisacariu]{bian2022nopenerf}
Wenjing Bian, Zirui Wang, Kejie Li, Jiawang Bian, and Victor~Adrian Prisacariu.
\newblock Nope-nerf: Optimising neural radiance field with no pose prior.
\newblock 2023.

\bibitem[Chen et~al.(2021)Chen, Xu, Zhao, Zhang, Xiang, Yu, and Su]{chen2021mvsnerf}
Anpei Chen, Zexiang Xu, Fuqiang Zhao, Xiaoshuai Zhang, Fanbo Xiang, Jingyi Yu, and Hao Su.
\newblock Mvsnerf: Fast generalizable radiance field reconstruction from multi-view stereo.
\newblock In \emph{Int. Conf. Comput. Vis.}, pages 14124--14133, 2021.

\bibitem[Chen et~al.(2022)Chen, Xu, Geiger, Yu, and Su]{chen2022tensorf}
Anpei Chen, Zexiang Xu, Andreas Geiger, Jingyi Yu, and Hao Su.
\newblock Tensorf: Tensorial radiance fields.
\newblock In \emph{Eur. Conf. Comput. Vis.}, pages 333--350. Springer, 2022.

\bibitem[Chen et~al.(2024{\natexlab{a}})Chen, Li, Ye, Wang, Xie, Zhai, Wang, Liu, Bao, and Zhang]{chen2024pgsr}
Danpeng Chen, Hai Li, Weicai Ye, Yifan Wang, Weijian Xie, Shangjin Zhai, Nan Wang, Haomin Liu, Hujun Bao, and Guofeng Zhang.
\newblock Pgsr: Planar-based gaussian splatting for efficient and high-fidelity surface reconstruction.
\newblock \emph{IEEE Transactions on Visualization and Computer Graphics}, 2024{\natexlab{a}}.

\bibitem[Chen et~al.(2023)Chen, Li, and Lee]{chen2023neusg}
Hanlin Chen, Chen Li, and Gim~Hee Lee.
\newblock Neusg: Neural implicit surface reconstruction with 3d gaussian splatting guidance.
\newblock \emph{arXiv preprint arXiv:2312.00846}, 2023.

\bibitem[Chen et~al.(2024{\natexlab{b}})Chen, He, Huang, Ye, Chen, Tang, Chen, Cai, Yang, Yu, et~al.]{chen2024meshanything}
Yiwen Chen, Tong He, Di Huang, Weicai Ye, Sijin Chen, Jiaxiang Tang, Xin Chen, Zhongang Cai, Lei Yang, Gang Yu, et~al.
\newblock Meshanything: Artist-created mesh generation with autoregressive transformers.
\newblock \emph{arXiv preprint arXiv:2406.10163}, 2024{\natexlab{b}}.

\bibitem[De~Boor(1978)]{de1978splines}
C De~Boor.
\newblock A practical guide to splines.
\newblock \emph{Springer-Verlag google schola}, 2:\penalty0 4135--4195, 1978.

\bibitem[Fridovich-Keil et~al.(2022)Fridovich-Keil, Yu, Tancik, Chen, Recht, and Kanazawa]{fridovich2022plenoxels}
Sara Fridovich-Keil, Alex Yu, Matthew Tancik, Qinhong Chen, Benjamin Recht, and Angjoo Kanazawa.
\newblock Plenoxels: Radiance fields without neural networks.
\newblock In \emph{IEEE Conf. Comput. Vis. Pattern Recog.}, pages 5501--5510, 2022.

\bibitem[Gu{\'e}don and Lepetit(2024)]{guedon2024sugar}
Antoine Gu{\'e}don and Vincent Lepetit.
\newblock Sugar: Surface-aligned gaussian splatting for efficient 3d mesh reconstruction and high-quality mesh rendering.
\newblock In \emph{IEEE Conf. Comput. Vis. Pattern Recog.}, pages 5354--5363, 2024.

\bibitem[He et~al.(2024)He, Osman, and Chaudhari]{he2024nerfsback}
Siming He, Zach Osman, and Pratik Chaudhari.
\newblock From nerfs to gaussian splats, and back.
\newblock \emph{arXiv preprint arXiv:2405.09717}, 2024.

\bibitem[Huang et~al.(2024)Huang, Yu, Chen, Geiger, and Gao]{Huang2DGS2024}
Binbin Huang, Zehao Yu, Anpei Chen, Andreas Geiger, and Shenghua Gao.
\newblock 2d gaussian splatting for geometrically accurate radiance fields.
\newblock In \emph{SIGGRAPH 2024 Conference Papers}. Association for Computing Machinery, 2024.

\bibitem[Jensen et~al.(2014)Jensen, Dahl, Vogiatzis, Tola, and Aanæs]{DTU2014}
Rasmus Jensen, Anders Dahl, George Vogiatzis, Engil Tola, and Henrik Aanæs.
\newblock Large scale multi-view stereopsis evaluation.
\newblock In \emph{IEEE Conf. Comput. Vis. Pattern Recog.}, 2014.

\bibitem[Kerbl et~al.(2023)Kerbl, Kopanas, Leimk{\"u}hler, and Drettakis]{kerbl20233d}
Bernhard Kerbl, Georgios Kopanas, Thomas Leimk{\"u}hler, and George Drettakis.
\newblock 3d gaussian splatting for real-time radiance field rendering.
\newblock \emph{ACM Trans. Graph.}, 2023.

\bibitem[Kim and Lim(2024)]{kim2024integrating}
Jiyeop Kim and Jongwoo Lim.
\newblock Integrating meshes and 3d gaussians for indoor scene reconstruction with sam mask guidance.
\newblock \emph{arXiv preprint arXiv:2407.16173}, 2024.

\bibitem[Li et~al.(2023{\natexlab{a}})Li, Bian, Castle, Torr, and Prisacariu]{li2023mobilebrick}
Kejie Li, Jia-Wang Bian, Robert Castle, Philip~HS Torr, and Victor~Adrian Prisacariu.
\newblock Mobilebrick: Building lego for 3d reconstruction on mobile devices.
\newblock In \emph{IEEE Conf. Comput. Vis. Pattern Recog.}, pages 4892--4901, 2023{\natexlab{a}}.

\bibitem[Li et~al.(2023{\natexlab{b}})Li, M{\"u}ller, Evans, Taylor, Unberath, Liu, and Lin]{li2023neuralangelo}
Zhaoshuo Li, Thomas M{\"u}ller, Alex Evans, Russell~H Taylor, Mathias Unberath, Ming-Yu Liu, and Chen-Hsuan Lin.
\newblock Neuralangelo: High-fidelity neural surface reconstruction.
\newblock In \emph{IEEE Conf. Comput. Vis. Pattern Recog.}, pages 8456--8465, 2023{\natexlab{b}}.

\bibitem[Liang et~al.(2023)Liang, He, and Chen]{liang2023retr}
Yixun Liang, Hao He, and Yingcong Chen.
\newblock Retr: Modeling rendering via transformer for generalizable neural surface reconstruction.
\newblock \emph{Advances in Neural Information Processing Systems}, 36:\penalty0 62332--62351, 2023.

\bibitem[Long et~al.(2022)Long, Lin, Wang, Komura, and Wang]{long2022sparseneus}
Xiaoxiao Long, Cheng Lin, Peng Wang, Taku Komura, and Wenping Wang.
\newblock Sparseneus: Fast generalizable neural surface reconstruction from sparse views.
\newblock In \emph{Eur. Conf. Comput. Vis.}, pages 210--227. Springer, 2022.

\bibitem[Lorensen and Cline(1998)]{lorensen1998marching}
William~E Lorensen and Harvey~E Cline.
\newblock Marching cubes: A high resolution 3d surface construction algorithm.
\newblock In \emph{Seminal graphics: pioneering efforts that shaped the field}, pages 347--353, 1998.

\bibitem[Lu et~al.(2024)Lu, Yu, Xu, Xiangli, Wang, Lin, and Dai]{lu2024scaffold}
Tao Lu, Mulin Yu, Linning Xu, Yuanbo Xiangli, Limin Wang, Dahua Lin, and Bo Dai.
\newblock Scaffold-gs: Structured 3d gaussians for view-adaptive rendering.
\newblock In \emph{IEEE Conf. Comput. Vis. Pattern Recog.}, pages 20654--20664, 2024.

\bibitem[Lyu et~al.(2024)Lyu, Sun, Huang, Wu, Yang, Chen, Pang, and Qi]{lyu20243dgsr}
Xiaoyang Lyu, Yang-Tian Sun, Yi-Hua Huang, Xiuzhe Wu, Ziyi Yang, Yilun Chen, Jiangmiao Pang, and Xiaojuan Qi.
\newblock 3dgsr: Implicit surface reconstruction with 3d gaussian splatting.
\newblock \emph{ACM Transactions on Graphics (TOG)}, 43\penalty0 (6):\penalty0 1--12, 2024.

\bibitem[Mildenhall et~al.(2020)Mildenhall, Srinivasan, Tancik, Barron, Ramamoorthi, and Ng]{mildenhall2020nerf}
Ben Mildenhall, Pratul~P. Srinivasan, Matthew Tancik, Jonathan~T. Barron, Ravi Ramamoorthi, and Ren Ng.
\newblock Nerf: Representing scenes as neural radiance fields for view synthesis.
\newblock In \emph{Eur. Conf. Comput. Vis.}, 2020.

\bibitem[M{\"u}ller et~al.(2022)M{\"u}ller, Evans, Schied, and Keller]{muller2022instant}
Thomas M{\"u}ller, Alex Evans, Christoph Schied, and Alexander Keller.
\newblock Instant neural graphics primitives with a multiresolution hash encoding.
\newblock \emph{ACM Trans. Graph.}, 41\penalty0 (4):\penalty0 1--15, 2022.

\bibitem[Osher et~al.(2003)Osher, Fedkiw, Osher, and Fedkiw]{osher2003constructing}
Stanley Osher, Ronald Fedkiw, Stanley Osher, and Ronald Fedkiw.
\newblock Constructing signed distance functions.
\newblock \emph{Level set methods and dynamic implicit surfaces}, pages 63--74, 2003.

\bibitem[Park et~al.(2019)Park, Florence, Straub, Newcombe, and Lovegrove]{park2019deepsdf}
Jeong~Joon Park, Peter Florence, Julian Straub, Richard Newcombe, and Steven Lovegrove.
\newblock Deepsdf: Learning continuous signed distance functions for shape representation.
\newblock In \emph{IEEE Conf. Comput. Vis. Pattern Recog.}, pages 165--174, 2019.

\bibitem[Ren et~al.(2023)Ren, Wang, Zhang, Pollefeys, and S{\"u}sstrunk]{ren2023volrecon}
Yufan Ren, Fangjinhua Wang, Tong Zhang, Marc Pollefeys, and Sabine S{\"u}sstrunk.
\newblock Volrecon: Volume rendering of signed ray distance functions for generalizable multi-view reconstruction.
\newblock In \emph{IEEE Conf. Comput. Vis. Pattern Recog.}, pages 16685--16695, 2023.

\bibitem[Sch\"{o}nberger and Frahm(2016)]{schoenberger2016sfm}
Johannes~Lutz Sch\"{o}nberger and Jan-Michael Frahm.
\newblock Structure-from-motion revisited.
\newblock In \emph{IEEE Conf. Comput. Vis. Pattern Recog.}, 2016.

\bibitem[Sch{\"o}nberger et~al.(2016)Sch{\"o}nberger, Zheng, Frahm, and Pollefeys]{schonberger2016colmap}
Johannes~L Sch{\"o}nberger, Enliang Zheng, Jan-Michael Frahm, and Marc Pollefeys.
\newblock Pixelwise view selection for unstructured multi-view stereo.
\newblock In \emph{Eur. Conf. Comput. Vis.}, pages 501--518. Springer, 2016.

\bibitem[Snavely et~al.(2006)Snavely, Seitz, and Szeliski]{snavely2006sfm}
Noah Snavely, Steven~M Seitz, and Richard Szeliski.
\newblock Photo tourism: exploring photo collections in 3d.
\newblock In \emph{ACM Siggraph}, pages 835--846, 2006.

\bibitem[Song et~al.(2024)Song, Zheng, Yuan, Gao, Zhao, He, Gu, and Zhao]{song2024sa}
Xiaowei Song, Jv Zheng, Shiran Yuan, Huan-ang Gao, Jingwei Zhao, Xiang He, Weihao Gu, and Hao Zhao.
\newblock Sa-gs: Scale-adaptive gaussian splatting for training-free anti-aliasing.
\newblock \emph{arXiv preprint arXiv:2403.19615}, 2024.

\bibitem[Sun et~al.(2022)Sun, Sun, and Chen]{sun2022direct}
Cheng Sun, Min Sun, and Hwann-Tzong Chen.
\newblock Direct voxel grid optimization: Super-fast convergence for radiance fields reconstruction.
\newblock In \emph{IEEE Conf. Comput. Vis. Pattern Recog.}, pages 5459--5469, 2022.

\bibitem[Sun et~al.(2024)Sun, Qin, Wang, Yan, Liu, Jia, and Shi]{sun20243dgs}
Hao Sun, Junping Qin, Lei Wang, Kai Yan, Zheng Liu, Xinglong Jia, and Xiaole Shi.
\newblock 3dgs-hd: Elimination of unrealistic artifacts in 3d gaussian splatting.
\newblock In \emph{2024 6th International Conference on Data-driven Optimization of Complex Systems (DOCS)}, pages 696--702. IEEE, 2024.

\bibitem[Turkulainen et~al.(2025)Turkulainen, Ren, Melekhov, Seiskari, Rahtu, and Kannala]{turkulainen2024dn}
Matias Turkulainen, Xuqian Ren, Iaroslav Melekhov, Otto Seiskari, Esa Rahtu, and Juho Kannala.
\newblock Dn-splatter: Depth and normal priors for gaussian splatting and meshing.
\newblock In \emph{2025 IEEE/CVF Winter Conference on Applications of Computer Vision (WACV)}, pages 2421--2431. IEEE, 2025.

\bibitem[Wang et~al.(2021)Wang, Liu, Liu, Theobalt, Komura, and Wang]{wang2021neus}
Peng Wang, Lingjie Liu, Yuan Liu, Christian Theobalt, Taku Komura, and Wenping Wang.
\newblock Neus: Learning neural implicit surfaces by volume rendering for multi-view reconstruction.
\newblock In \emph{Adv. Neural Inform. Process. Syst.}, 2021.

\bibitem[Wang et~al.(2024)Wang, Hua, Shingys, Niu, Yang, Gao, Zheng, Yang, and Wang]{wang2024enhancement}
Ruizhe Wang, Chunliang Hua, Tomakayev Shingys, Mengyuan Niu, Qingxin Yang, Lizhong Gao, Yi Zheng, Junyan Yang, and Qiao Wang.
\newblock Enhancement of 3d gaussian splatting using raw mesh for photorealistic recreation of architectures.
\newblock \emph{arXiv preprint arXiv:2407.15435}, 2024.

\bibitem[WE(1987)]{we1987marching}
LORENSEN WE.
\newblock Marching cubes: A high resolution 3d surface construction algorithm.
\newblock \emph{Computer graphics}, 21\penalty0 (1):\penalty0 7--12, 1987.

\bibitem[Wei et~al.(2024)Wei, Zhang, Bi, Tan, Luan, Deschaintre, Sunkavalli, Su, and Xu]{wei2024meshlrm}
Xinyue Wei, Kai Zhang, Sai Bi, Hao Tan, Fujun Luan, Valentin Deschaintre, Kalyan Sunkavalli, Hao Su, and Zexiang Xu.
\newblock Meshlrm: Large reconstruction model for high-quality meshes.
\newblock \emph{arXiv preprint arXiv:2404.12385}, 2024.

\bibitem[Wu et~al.(2024)Wu, Lin, Zhang, Zeng, Xu, Torr, Cao, and Yao]{wu2024direct3d}
Shuang Wu, Youtian Lin, Feihu Zhang, Yifei Zeng, Jingxi Xu, Philip Torr, Xun Cao, and Yao Yao.
\newblock Direct3d: Scalable image-to-3d generation via 3d latent diffusion transformer.
\newblock \emph{Advances in Neural Information Processing Systems}, 37:\penalty0 121859--121881, 2024.

\bibitem[Wu et~al.(2023)Wu, Wang, Pan, Xu, Theobalt, Liu, and Lin]{wu2022voxurf}
Tong Wu, Jiaqi Wang, Xingang Pan, Xudong Xu, Christian Theobalt, Ziwei Liu, and Dahua Lin.
\newblock Voxurf: Voxel-based efficient and accurate neural surface reconstruction.
\newblock In \emph{Int. Conf. Learn. Represent.}, 2023.

\bibitem[Xiong et~al.(2023)Xiong, Muttukuru, Upadhyay, Chari, and Kadambi]{xiong2023sparsegs}
Haolin Xiong, Sairisheek Muttukuru, Rishi Upadhyay, Pradyumna Chari, and Achuta Kadambi.
\newblock Sparsegs: Real-time 360° sparse view synthesis using gaussian splatting.
\newblock \emph{Arxiv}, 2023.

\bibitem[Xu et~al.(2024)Xu, Cheng, Gao, Wang, Gao, and Shan]{xu2024instantmesh}
Jiale Xu, Weihao Cheng, Yiming Gao, Xintao Wang, Shenghua Gao, and Ying Shan.
\newblock Instantmesh: Efficient 3d mesh generation from a single image with sparse-view large reconstruction models.
\newblock \emph{arXiv preprint arXiv:2404.07191}, 2024.

\bibitem[Yang et~al.(2024{\natexlab{a}})Yang, Zhu, Jiang, Ye, Chen, Zhang, Chen, Zhao, and Zhao]{yang2024spectrally}
Runyi Yang, Zhenxin Zhu, Zhou Jiang, Baijun Ye, Xiaoxue Chen, Yifei Zhang, Yuantao Chen, Jian Zhao, and Hao Zhao.
\newblock Spectrally pruned gaussian fields with neural compensation.
\newblock \emph{arXiv preprint arXiv:2405.00676}, 2024{\natexlab{a}}.

\bibitem[Yang et~al.(2024{\natexlab{b}})Yang, Gao, Zhou, Jiao, Zhang, and Jin]{yang2024deformable}
Ziyi Yang, Xinyu Gao, Wen Zhou, Shaohui Jiao, Yuqing Zhang, and Xiaogang Jin.
\newblock Deformable 3d gaussians for high-fidelity monocular dynamic scene reconstruction.
\newblock In \emph{IEEE Conf. Comput. Vis. Pattern Recog.}, pages 20331--20341, 2024{\natexlab{b}}.

\bibitem[Yao et~al.(2020)Yao, Luo, Li, Zhang, Ren, Zhou, Fang, and Quan]{yao2020blendedmvs}
Yao Yao, Zixin Luo, Shiwei Li, Jingyang Zhang, Yufan Ren, Lei Zhou, Tian Fang, and Long Quan.
\newblock Blendedmvs: A large-scale dataset for generalized multi-view stereo networks.
\newblock In \emph{IEEE Conf. Comput. Vis. Pattern Recog.}, pages 1790--1799, 2020.

\bibitem[Yariv et~al.(2021)Yariv, Gu, Kasten, and Lipman]{yariv2021volsdf}
Lior Yariv, Jiatao Gu, Yoni Kasten, and Yaron Lipman.
\newblock Volume rendering of neural implicit surfaces.
\newblock \emph{Adv. Neural Inform. Process. Syst.}, 34:\penalty0 4805--4815, 2021.

\bibitem[Ye et~al.(2025)Ye, Li, Kerr, Turkulainen, Yi, Pan, Seiskari, Ye, Hu, Tancik, et~al.]{ye2024gsplat}
Vickie Ye, Ruilong Li, Justin Kerr, Matias Turkulainen, Brent Yi, Zhuoyang Pan, Otto Seiskari, Jianbo Ye, Jeffrey Hu, Matthew Tancik, et~al.
\newblock gsplat: An open-source library for gaussian splatting.
\newblock \emph{Journal of Machine Learning Research}, 26\penalty0 (34):\penalty0 1--17, 2025.

\bibitem[Yin et~al.(2023)Yin, Zhang, Chen, Cai, Yu, Wang, Chen, and Shen]{yin2023metric3d}
Wei Yin, Chi Zhang, Hao Chen, Zhipeng Cai, Gang Yu, Kaixuan Wang, Xiaozhi Chen, and Chunhua Shen.
\newblock Metric3d: Towards zero-shot metric 3d prediction from a single image.
\newblock In \emph{Int. Conf. Comput. Vis.}, pages 9043--9053, 2023.

\bibitem[Younes et~al.(2024)Younes, Ouasfi, and Boukhayma]{younes2024sparsecraft}
Mae Younes, Amine Ouasfi, and Adnane Boukhayma.
\newblock Sparsecraft: Few-shot neural reconstruction through stereopsis guided geometric linearization.
\newblock In \emph{European Conference on Computer Vision}, pages 37--56. Springer, 2024.

\bibitem[Yu et~al.(2021)Yu, Ye, Tancik, and Kanazawa]{yu2021pixelnerf}
Alex Yu, Vickie Ye, Matthew Tancik, and Angjoo Kanazawa.
\newblock pixelnerf: Neural radiance fields from one or few images.
\newblock In \emph{IEEE Conf. Comput. Vis. Pattern Recog.}, pages 4578--4587, 2021.

\bibitem[Yu et~al.(2024{\natexlab{a}})Yu, Lu, Xu, Jiang, Xiangli, and Dai]{yu2024gsdf}
Mulin Yu, Tao Lu, Linning Xu, Lihan Jiang, Yuanbo Xiangli, and Bo Dai.
\newblock Gsdf: 3dgs meets sdf for improved neural rendering and reconstruction.
\newblock \emph{Advances in Neural Information Processing Systems}, 37:\penalty0 129507--129530, 2024{\natexlab{a}}.

\bibitem[Yu et~al.(2022)Yu, Peng, Niemeyer, Sattler, and Geiger]{Yu2022MonoSDF}
Zehao Yu, Songyou Peng, Michael Niemeyer, Torsten Sattler, and Andreas Geiger.
\newblock Monosdf: Exploring monocular geometric cues for neural implicit surface reconstruction.
\newblock \emph{Adv. Neural Inform. Process. Syst.}, 2022.

\bibitem[Yu et~al.(2024{\natexlab{b}})Yu, Chen, Huang, Sattler, and Geiger]{yu2024mip}
Zehao Yu, Anpei Chen, Binbin Huang, Torsten Sattler, and Andreas Geiger.
\newblock Mip-splatting: Alias-free 3d gaussian splatting.
\newblock In \emph{IEEE Conf. Comput. Vis. Pattern Recog.}, pages 19447--19456, 2024{\natexlab{b}}.

\bibitem[Yu et~al.(2024{\natexlab{c}})Yu, Sattler, and Geiger]{Yu2024GOF}
Zehao Yu, Torsten Sattler, and Andreas Geiger.
\newblock Gaussian opacity fields: Efficient adaptive surface reconstruction in unbounded scenes.
\newblock \emph{ACM Trans. Graph.}, 2024{\natexlab{c}}.

\bibitem[Zhang et~al.(2024)Zhang, Fang, Shrestha, Liang, Long, and Tan]{zhang2024rade}
Baowen Zhang, Chuan Fang, Rakesh Shrestha, Yixun Liang, Xiaoxiao Long, and Ping Tan.
\newblock Rade-gs: Rasterizing depth in gaussian splatting.
\newblock \emph{arXiv preprint arXiv:2406.01467}, 2024.

\bibitem[Zhang et~al.(2020)Zhang, Riegler, Snavely, and Koltun]{zhang2020nerf++}
Kai Zhang, Gernot Riegler, Noah Snavely, and Vladlen Koltun.
\newblock Nerf++: Analyzing and improving neural radiance fields.
\newblock \emph{arXiv preprint arXiv:2010.07492}, 2020.

\end{thebibliography}
}

\clearpage
\setcounter{page}{1}
\maketitlesupplementary

\setlength{\tabcolsep}{3pt}

% The supplementary document is organized as follows:

% \secref{sec:A}: \textbf{Implementation Details} provides additional details on the SDF and 3DGS training process, including mesh sampling, training details and camera position settings.

% \secref{sec:B}: \textbf{More Experimental Results } shows the impact of sparsity on reconstruction accuracy, results for each scene on the MobileBrick dataset, reconstruction results from three input images on the DTU dataset, details on how SDF and 3DGS enhance each other, and reconstruction comparisons for three images on the BlendedMVS dataset.

% \secref{sec:C}: \textbf{More Qualitative Results} includes additional visual results that could not be displayed in the main text, mainly showcasing surface reconstruction on the DTU and MobileBrick datasets, as well as novel view renderings.

\appendix
\section{Implementation Details}
\label{sec:A}
% \subsection{Configurations}
% \paragraph{A complete cycle schedule.}
% During coarse mesh training, the voxel grid resolution is set to \(96^3\), with $10k$ iterations taking approximately 15 minutes. In coarse mesh training, a normal branch is added with a loss weight of $0.05$. All other parameter settings follow Voxurf~\cite{wu2022voxurf}. Subsequently, we train 3DGS~\cite{kerbl20233d} for $7k$ iterations, which requires around 5 minutes. Rendering 10 new viewpoint images takes only about 30 seconds. Once the Expanded Image Collection is obtained, the number of training images increases, prompting an increase in voxel grid resolution to \(256^3\) and $20k$ iterations, which takes approximately 40 minutes. Thus, a complete cycle requires roughly 1 hour.

\paragraph{Mesh sampling for 3DGS.}
In the main text Sec. 3.2, we sample surface points from coarse mesh as 3DGS~\cite{sun20243dgs} initialization. Specifically, we render depth maps of training viewpoints using the coarse mesh and sample $5k$ points from each depth map. Then, we unproject the depth points into 3D color points. These points are then fused to generate a total of $50k$ points, which are subsequently combined with the sparse results from COLMAP~\cite{schonberger2016colmap}.
\paragraph{3DGS training details.}
For 3DGS~\cite{sun20243dgs} training, convergence is achieved effectively with $7k$  iterations under sparse inputs. Specifically, densification begins at $500$ iterations with intervals of $100$ iterations. An opacity reset is performed at $3k$ iterations, while other parameters remain consistent with the original implementation. To ensure a fair comparison, other GS-based methods tested in the paper also follow this training strategy.

\paragraph{Camera position.}
Camera pose refers to the camera's position $\boldsymbol{c}$ and orientation matrix $\mathbf{R}$ in the world coordinate system.
The proposed two methods in the main text Sec. 3.3 (Camera position perturbation and interpolation) focus exclusively on resampling camera positions while ensuring that the camera orientation consistently points toward the center of the object, located at $(0,0,0)$ in the world coordinate system.

\section{More Experimental Results}
\label{sec:B}

\subsection{Results on different sparsity levels.}
To better understand the strengths and weaknesses of the SDF-based and 3DGS-based methods, we evaluate them under varying levels of sparse input. Specifically, we select 9 scenes from the MobileBrick dataset~\cite{li2023mobilebrick}, and the reported results are averaged across all scenes.
\tabref{tab:input_number} presents results across different sparsity levels. 3DGS-based methods (e.g., GOF\cite{Yu2024GOF}) significantly outperform SDF-based methods (e.g., Voxurf\cite{wu2022voxurf}) in novel view rendering, while Voxurf consistently achieves better surface reconstruction than GOF. We hypothesize that this stems from SDF’s dense representations, which effectively capture global geometry, and 3DGS’s sparse representations, which excel at preserving local details.
To leverage the strengths of both approaches, we propose a hybrid method, leading to our proposed \name framework.

\begin{table}[htbp]
  \centering
  \small
  \caption{Rendering and mesh reconstruction results on SDF-based and GS-based methods with different input image numbers.}
    \begin{tabular}{c|cc|cc}
    \toprule
          & \multicolumn{2}{c|}{Rendering (PSNR)} & \multicolumn{2}{c}{Mesh (F1 Score)} \\
\cmidrule{2-5}    Input & Voxurf\cite{wu2022voxurf} & GOF\cite{Yu2024GOF}   & Voxurf\cite{wu2022voxurf} & GOF\cite{Yu2024GOF} \\
    \midrule
    5     & 11.78  & 12.61  & 31.70  & 30.15  \\
    10    & 14.06  & 16.00  & 63.60  & 53.64  \\
    15    & 14.90  & 18.30  & 66.82  & 60.20  \\
    20    & 15.83  & 19.81  & 70.39  & 65.86  \\
    30    & 16.93  & 21.43  & 71.97  & 68.25  \\
    \bottomrule
    \end{tabular}%
  \label{tab:input_number}%
\end{table}%

% \begin{figure}
%   \centering
%   \includegraphics[width = 0.5\textwidth]{figure/sparsity.pdf}
%   \caption{Novel view rendering and mesh reconstruction quality with varying numbers of input images}
%   \label{fig:3dgs_psnr}
% \end{figure}

\begin{figure*}
  \centering
  \includegraphics[width = 1.0\textwidth]{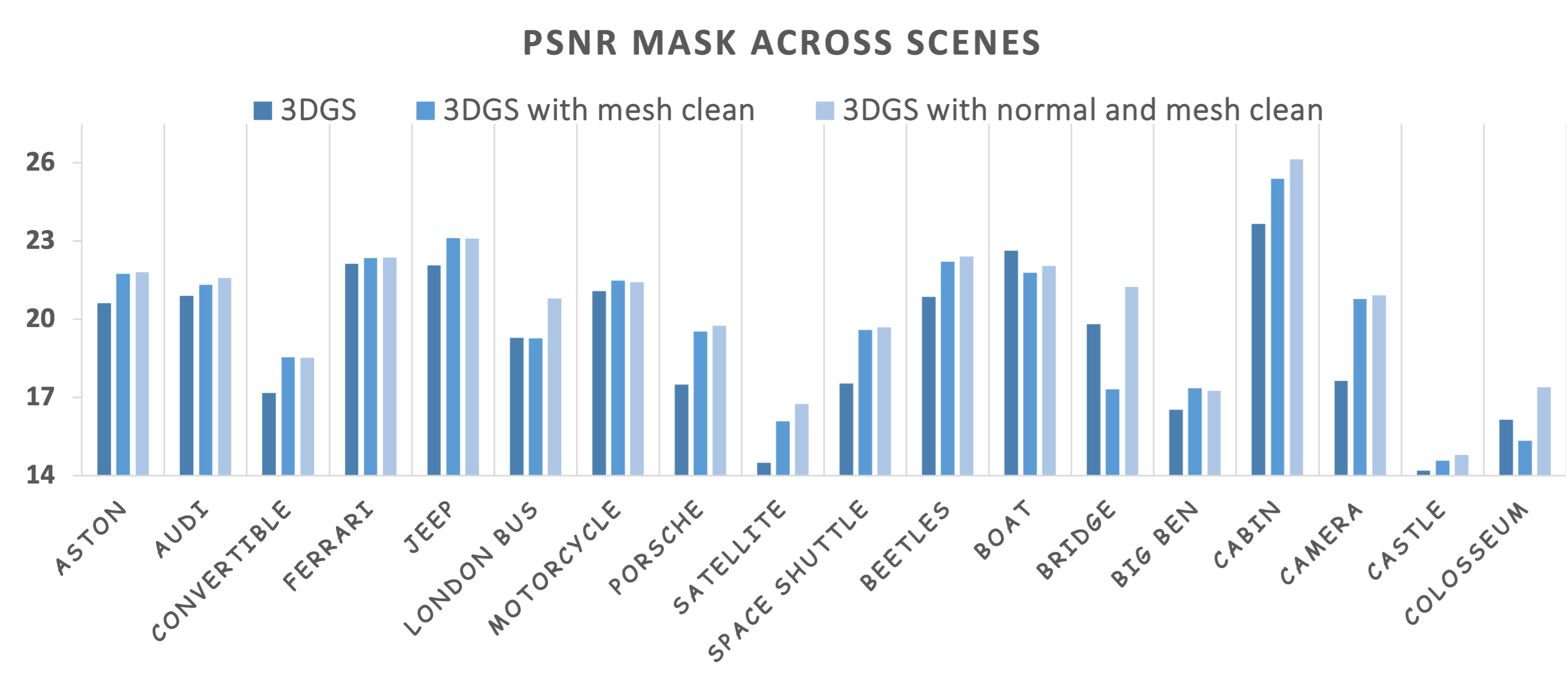}
  \caption{Ablation study on mesh-based sampling for enhancing 3DGS rendering. We report foreground PSNR here.}
  \label{fig:3dgs_psnr}
\end{figure*}

\begin{table*}[htbp]
  \centering
  \small
  \caption{Quantitative F1 score (↑) across all 18 MobileBrick test scenes.}
  \scalebox{0.75}{
    \begin{tabular}{ccccccccccccccccccccc}
    \toprule
    F1 Score & Aston & Conv. & Ferrari & Jeep  & Bus & Moto. & Porsche & Beetles & Big\_ben & Boat  & Audi  & Bridge & Cabin & Camera & Castle & Colosseum & Satellite & Shuttle & Mean & Time \\
    \midrule
    Voxurf~\cite{wu2022voxurf} & 55.8  & 53.1  & 69.4  & \textbf{88.0} & 58.7  & 88.2  & 59.7  & 64.8  & 55.0  & 60.6  & \textbf{83.1} & 67.9  & 78.4  & 91.0 & 11.5  & 22.3  & 61.7  & 60.9 & 62.9 & 55 mins \\
    \midrule
    MonoSDF~\cite{Yu2022MonoSDF} & 51.2  & 42.6  & 56.9  & 31.4  & 13.8  & 54.5  & 40.1  & 36.9  & 2.3  & 4.8  & 60.2  & 67.5  & 76.3  & 25.6  & 4.9  & 4.7  & 38.9  & 49.5 & 36.8 & 6 hrs \\
    \midrule
    2DGS~\cite{Huang2DGS2024}  & 42.8  & 36.5  & 62.1  & 71.7  & 34.9  & 51.6  & 39.3  & 47.2  & 28.4  & 67.8  & 73.9  & 81.2  & 62.5  & 43.7  & 17.3 & 18.9  & 4.6  & 40.8 & 45.8 & 10 mins \\
    \midrule
    GOF~\cite{Yu2024GOF}   & 55.7  & 48.3  & 67.8  & 70.2  & 46.4  & 73.5  & 50.9  & 62.7  & 42.3  & 71.9  & 77.2  & 78.6  & 73.8  & 53.4  & 13.6  & 26.5  & 33.8  & 50.2 & 55.4 & 50 mins \\
    \midrule
    Ours  & \textbf{60.8} & \textbf{58.9} & \textbf{70.1} & 86.5  & \textbf{67.3} & \textbf{89.8} & \textbf{62.4} & \textbf{76.4} & \textbf{62.3} & \textbf{81.5} & 80.9  & \textbf{94.3} & \textbf{80.6} & \textbf{91.4} & \textbf{17.8} & \textbf{32.1} & \textbf{73.5} & \textbf{64.7} & \textbf{69.0} & 1 hr \\
    \bottomrule
    \end{tabular}}%
  \label{tab:per_scene_mobilebrick}%
\end{table*}%

\begin{table*}[h]
  \centering
  \small
  \caption{Quantitative results of 3-view reconstruction on DTU. Chamfer Distance (mm)↓ is reported. 
  % The results of MonoSDF are reproduced by us. 
  Note that SparseNeuS requires pertaining on large-scale dataset and ground-truth masks at inference time. The best results are \textbf{bolded}, while the second-best are \underline{underlined}.}
  {\begin{tabular}{c|ccccccccccccccccc}
    \toprule
    Scan  & 24    & 37    & 40    & 55    & 63    & 65    & 69    & 83    & 97    & 105   & 106   & 110   & 114   & 118   & 122   & Mean  & Time\\
    \midrule
    Voxurf~\cite{wu2022voxurf} & 3.75  & 6.02  & 4.56  & 3.62  & 4.53  & 2.80  & 3.79  & 4.23  & 4.26  & 2.09  & 4.40  & 4.44  & 1.36   & 4.60    & 2.51  & 3.79 & 50 mins \\
    \midrule
    % MonoSDF~\cite{Yu2022MonoSDF} & \underline{3.47}  & \underline{3.61}  & 2.10  & \underline{1.05}  & 2.37  & \textbf{1.38}  & 1.41  &\underline{1.85}  & \underline{1.74}  & \underline{1.10}  & \textbf{1.46}  & 2.28  & 1.25   & \underline{1.44}  & \textbf{1.45}  & \textbf{1.68} & 6hr \\
    % \midrule
    MonoSDF~\cite{Yu2022MonoSDF} & 6.76  & 3.50  & \textbf{1.79}  & 0.73  & 1.95  & \textbf{1.45}  & \underline{1.25}  & 1.63  & \textbf{1.40} & \underline{0.98}  & 4.03  & \underline{1.75}  & 0.94   & 2.54  & 3.55  & 2.28 & 6 hrs \\
    \midrule
    SparseNeuS~\cite{long2022sparseneus} &     4.10   & 4.21  & 3.64  & 1.78  & 2.89  & 2.49  & 1.76  & 2.50   & 2.88  & 2.16  & 2.04  & 3.27  & 1.29  & 2.36  & 1.75  & 2.61 &Pretrain + 2 hrs ft \\
    \midrule
    VolRecon~\cite{ren2023volrecon} & 3.56  & 4.48  & 4.24  & 3.15  & 2.85  & 3.91  & 2.51  & 2.65  & 2.56  & 2.67  & 2.84  & 2.77  & 1.60  & 3.09  & 2.19  & 3.00  & 2days pre \\
    \midrule
    ReTR~\cite{liang2023retr} & 3.78  & 3.91  & 3.95  & 3.15  & 2.91  & 3.50  & 2.79  & 2.76  & 2.50  & 2.35  & 3.56  & 4.02  & 1.70  & 2.72  & 2.16  & 3.05  & 3days pre \\
    \midrule
    SparseCraft~\cite{younes2024sparsecraft} & \underline{2.13}  & 2.83  & 2.68  & \textbf{0.70} & 1.49  & 2.15  & 1.29  & \textbf{1.37} & 1.57  & 1.13  & \textbf{1.22} & 2.53  & \textbf{0.61} & \textbf{0.83} & \underline{0.99}  & \underline{1.57}  & 1.5 hours \\
    \midrule
    Ours(Voxurf) & 2.65  & 4.47  & 1.87 & 1.22  & \underline{2.28}  & \underline{1.98} & 1.33  & 1.96  & 2.66  & 1.94  & 1.86  & \textbf{1.67} & 0.78  & 1.22  & 1.63  & 1.96  & 1hour \\
    Ours(SparseCraft) & \textbf{1.86} & \textbf{2.56} & 2.85  & 0.75  & \textbf{1.40} & 1.99  & \textbf{1.13} & \underline{1.42}  & \underline{1.51} & \textbf{0.90} & \underline{1.28}  & 2.26  & \underline{0.68}  & \underline{0.89}  & \textbf{0.94} & \textbf{1.49} & 2 hours \\
    % SparseNeuS{womask} &     6.68   & 6.50  & 6.23  & 2.35  & 4.59  & 5.69  & 8.15  & 5.32   & 6.03  & 5.22  & 6.18  & 5.69  & 7.54  & 6.85  & 6.35  & 5.96 \\
    % \midrule
    % Ours  & \textbf{2.65}  & 4.47  & \underline{1.87}  & \underline{1.22}  & \underline{2.28}  & \underline{1.98}  & \underline{1.33}  & \underline{1.96}  & \underline{2.66}  & \underline{1.94} & \underline{1.86}  & \textbf{1.67}  & \textbf{0.78}  &\textbf{1.22}   & \textbf{1.63}  & \textbf{1.96} & 1 hr \\
    
    \bottomrule
    \end{tabular}}
  \label{table:DTU_results}%
\end{table*}%

\subsection{Per-scene 10-view mesh results on Mobilebrick}

\tabref{tab:per_scene_mobilebrick} presents the surface reconstruction results (F1 scores) for each MobileBrick scene, using 10 input images per scene for surface reconstruction. The best scores are highlighted in bold.

\subsection{Per-scene 3-view reconstruction mesh on DTU}
Previous methods~\cite{yu2021pixelnerf, Yu2022MonoSDF} use 3 manually selected images with the best overlap for surface reconstruction. However, we argue that this does not reflect real-world reconstruction scenarios. Instead, we propose evenly sampling 5 images for sparse-view reconstruction.
Nonetheless, we also report results under the 3-view setting for fair comparison with previous methods. \tabref{table:DTU_results} presents the results, demonstrating that our method outperforms existing alternatives. Furthermore, our framework is compatible with a variety of SDF-based methods and 3D Gaussian representations. In addition to integrating Voxurf into our pipeline, we also experiment with incorporating SparseCraft, and observe similarly strong reconstruction performance, demonstrating the generality and versatility of our approach.

\subsection{SDF-3DGS Mutual Enhancement.}
Our method enables mesh reconstruction and 3DGS to enhance each other’s performance. \tabref{tab:connectivity} presents the ablation study results on MobileBrick, demonstrating the effectiveness of this mutual enhancement. The results show that without support from the other module, performance drops significantly for both components.
Additionally, we analyze the impact of cyclic optimization in our method. Running two cycles provides a slight performance improvement. However, for a trade-off between efficiency and performance, we use a single loop iteration as the default setting.
\begin{table}[htbp]
  \centering
  \small
  \caption{Ablations studies on effectiveness of our proposed modules on MobileBrick test scenes.}
    \begin{tabular}{l|c|c|c|c}
    \toprule
          & \multicolumn{2}{c|}{Meshing} & \multicolumn{2}{c}{Rendering} \\
\cmidrule{1-5}  &  F1↑ & CD↓  & PSNR↑ & PSNR-F↑  \\
    \midrule
    SDF-based method w/o 3DGS & \textbf{62.42} & \textbf{13.3} & 14.34 & 18.34 \\
    3DGS-based method w/o SDF & 54.96  & 11.0 & \textbf{16.52} & \textbf{18.36} \\

\midrule
    Ours (One cycle) & 68.97 & \textbf{9.7} & 17.48 & 20.45 \\
    Ours (Two cycles) & \textbf{69.14} & 9.9 & \textbf{17.58} & \textbf{20.55} \\
    \bottomrule
    \end{tabular}%
  \label{tab:connectivity}%
\end{table}%

\subsection{Efficacy of mesh-based sampling for 3DGS}
Fig. 3~(e)\&(f) in main paper provide a visual comparison between our mesh-based point sampling approach and COLMAP-generated sparse points. The comparison shows that our method achieves noticeably better visual quality in object regions, which leads to enhanced 3DGS rendering quality. The results across each scene on MobileBrick are summarized in \figref{fig:3dgs_psnr}.
This demonstrates the effectiveness of our mesh cleaning and normal loss in enhancing 3DGS~\cite{sun20243dgs} rendering quality.

% \subsection{Training time}
% In \tabref{tab:addlabel}, we compared the training time of each method with $10$ input images, and our method achieved the best balance between time and performance.

\subsection{Efficacy of 3DGS for mesh reconstruction}
\begin{figure}
\centering
\includegraphics[width = 0.48\textwidth]{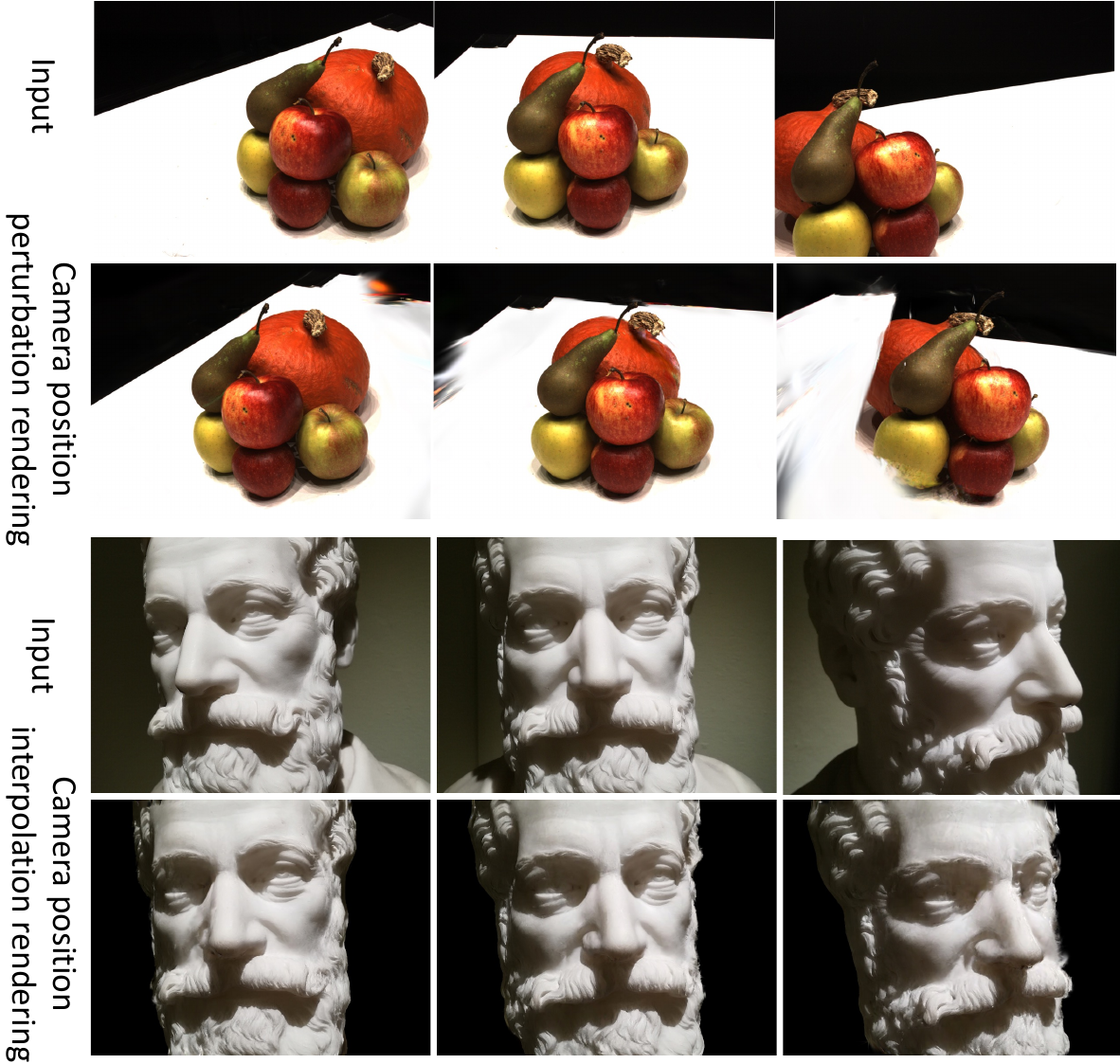}
  \caption{Visualization of newly rendered images with different pose expansion strategies. The top row presents results on DTU~\cite{DTU2014} (scan63), while the bottom row shows results on BlendedMVS~\cite{yao2020blendedmvs} (Man).
  }
  \label{fig:cam_pose_render_vis}
\end{figure}

Sec. 3.3 in the main text mentioned that 3DGS~\cite{sun20243dgs} can provide higher-quality novel view images, as extended views, are combined with the original inputs to refine the mesh. Specifically, we propose two novel view pose strategies, and we visualize the resulting novel view images in ~\figref{fig:cam_pose_render_vis}.

\subsection{Visual reconstruction on BlendedMVS}
\begin{figure}
  \centering
  \includegraphics[width = 0.5\textwidth]{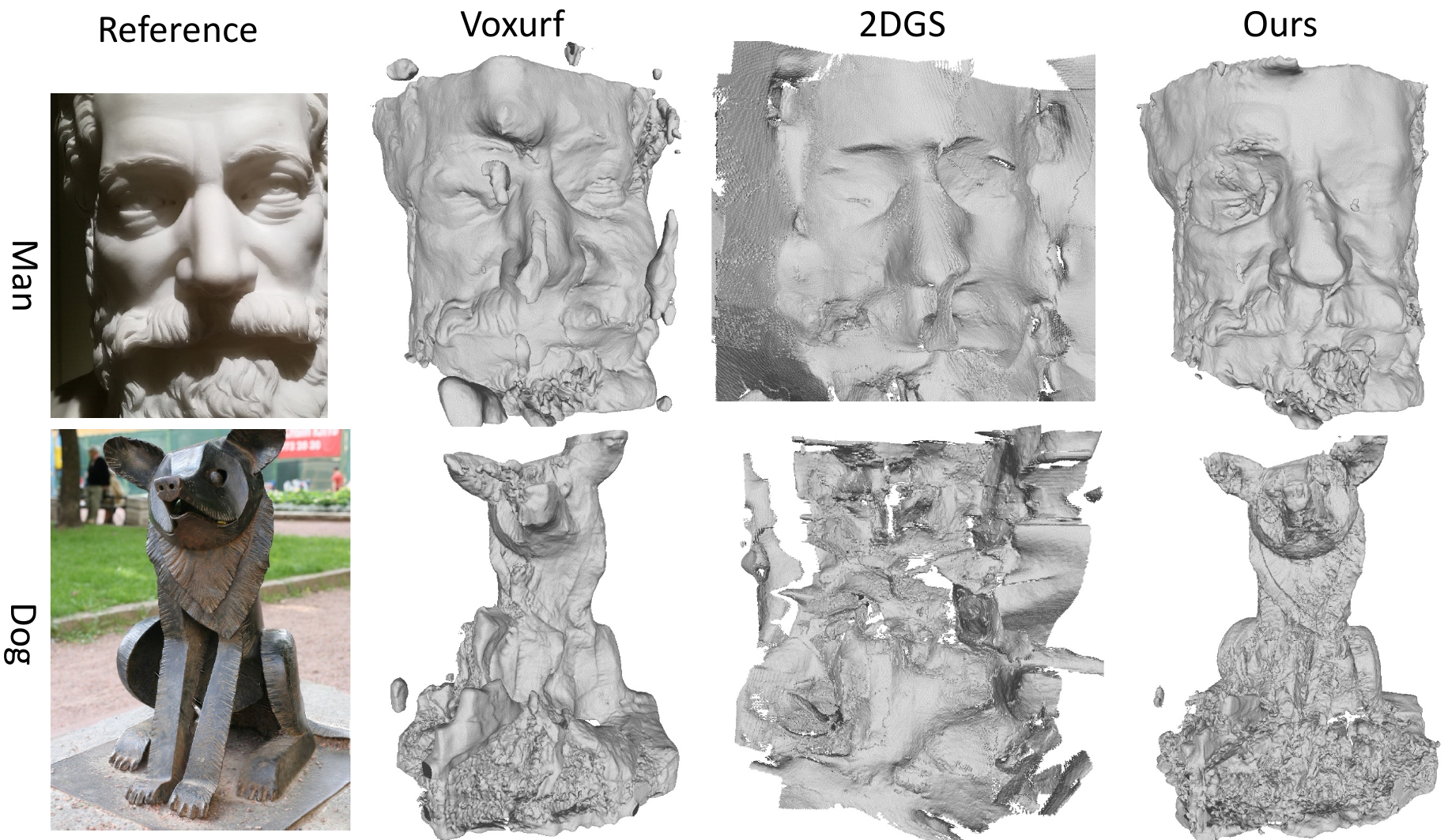}
  \caption{Qualitative comparison of 3-view mesh reconstruction on
BlendedMVS dataset.}
  \label{fig:bmvs_mesh}
\end{figure}

We performed mesh reconstruction using 3-view input on the BlendedMVS dataset~\cite{yao2020blendedmvs}. \figref{fig:bmvs_mesh} presents the results, comparing our method with two representative approaches: Voxurf (SDF-based) and 2DGS (3DGS-based).
While none of the methods perform well in this setting, our approach achieves slightly better results than the alternatives. We hypothesize that 3 input views are insufficient for real-world surface reconstruction, highlighting the challenges of extreme sparsity.

\section{More Qualitative Results}
\label{sec:C}

\subsection{DTU rendering results}
We visualize and compare the novel view synthesis results of our method (based on SparseCraft) against the original SparseCraft on the DTU dataset under sparse input settings of 3, 6, and 9 views.
\begin{figure}
\centering
\includegraphics[width = 0.5\textwidth]{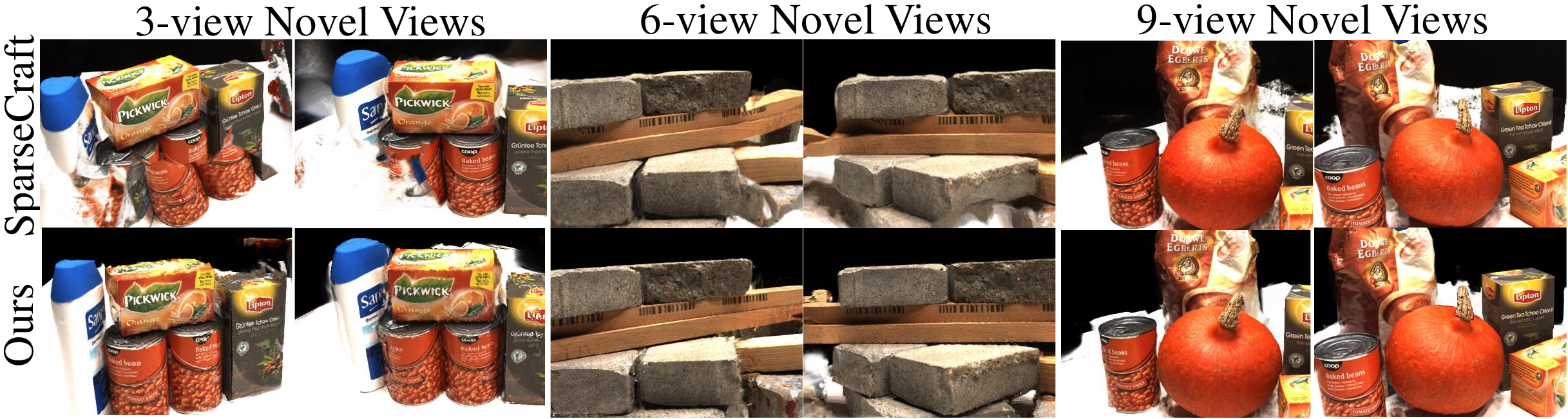}
  \caption{DTU novel view synthesis comparison.
  }
  \label{fig:dtu_3}
\end{figure}

\subsection{Mesh reconstruction on Mobilebrick and DTU}
\begin{figure*}
  \centering
  \includegraphics[width = 1.0\textwidth]{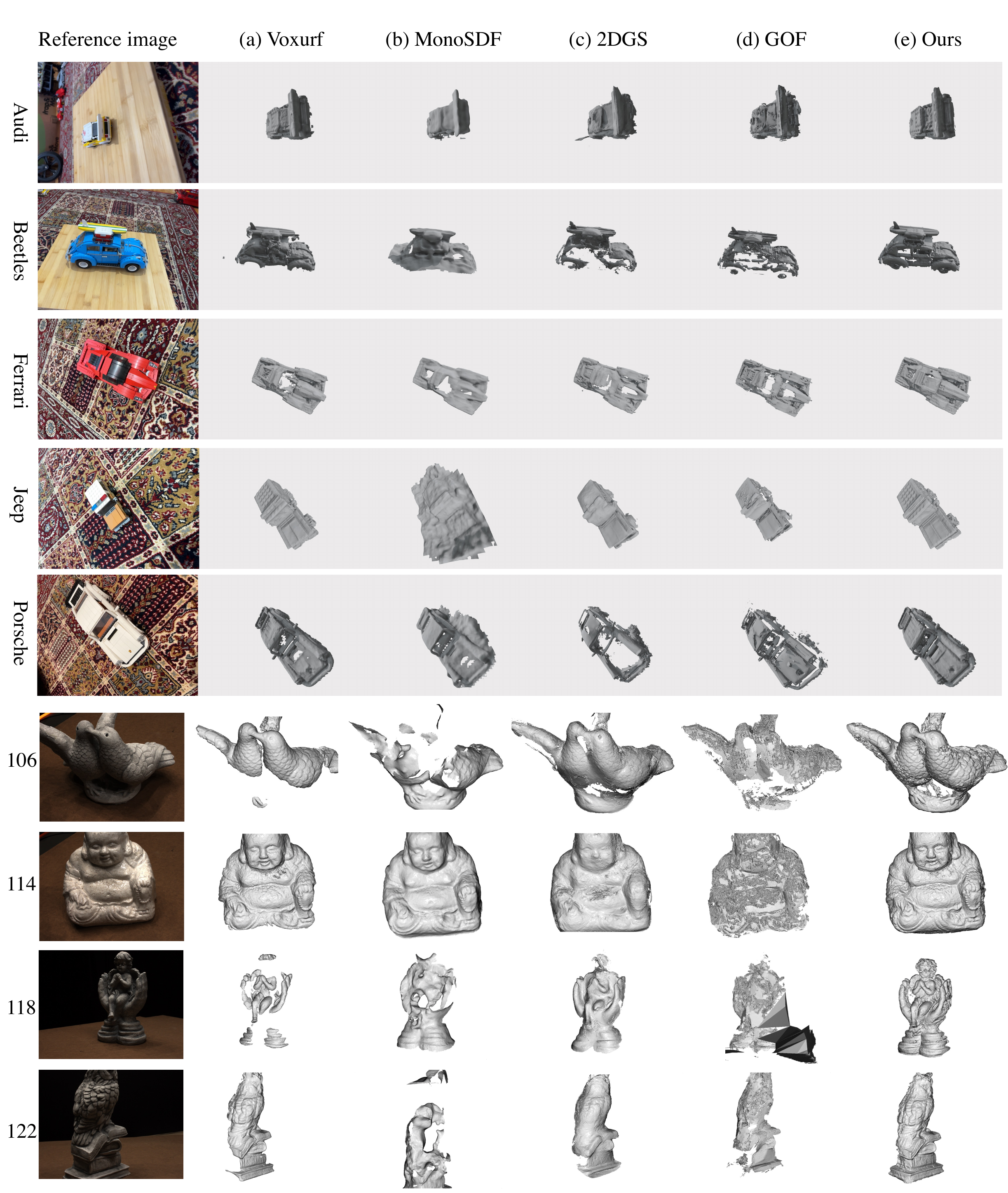}
  \caption{More qualitative mesh reconstruction results on MobileBrick and DTU.}
  \label{fig:vis_mesh_supp}
\end{figure*}

\figref{fig:vis_mesh_supp} Presents additional mesh reconstruction results on MobileBrick (10 images) and DTU (5 images). Training images are uniformly sampled to minimize overlap, making the task more challenging and reflect the real-world reconstruction problem.

% \subsection{Mesh reconstruction on DTU}
% \begin{figure*}
%   \centering
%   \includegraphics[width = 1.0\textwidth]{figure/summary_mesh_supp_dtu.pdf}
%   \caption{More qualitative mesh reconstruction results on DTU.}
%   \label{fig:vis_mesh_supp_dtu}
% \end{figure*}
\subsection{MobileBrick rendering results}
\begin{figure*}
  \centering
  \includegraphics[width = 0.9\textwidth]{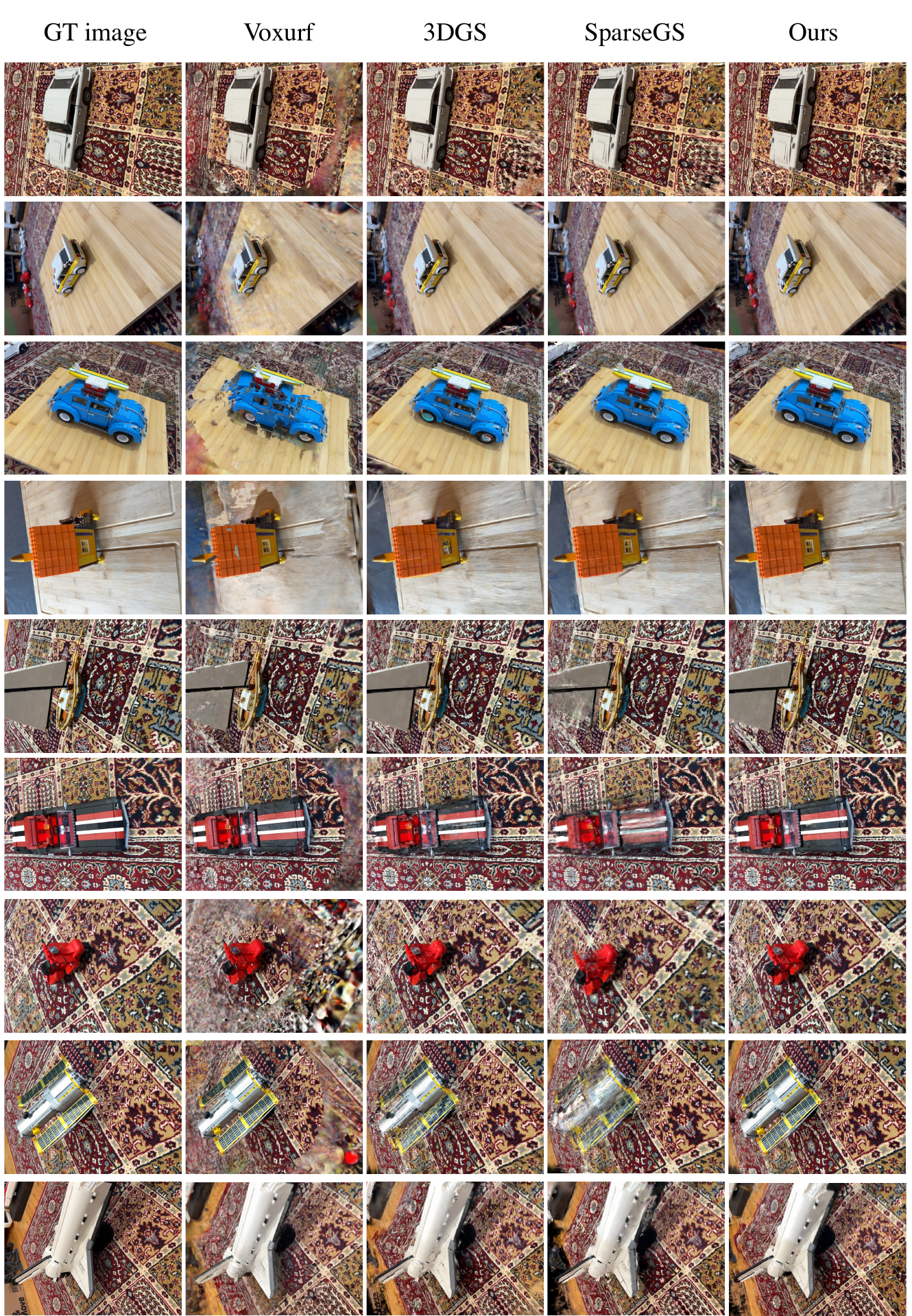}
  \caption{More qualitative novel view rendering results on MobileBrick.}
  \label{fig:vis_rendering_supp}
\end{figure*}

\figref{fig:vis_rendering_supp} presents additional novel view renderings on MobileBrick, demonstrating that our method achieves superior rendering quality. This improvement stems from the stable initialization point cloud provided by the coarse mesh.

% 
% {
%     \small
%     \bibliographystyle{ieeenat_fullname}
%     \bibliography{main}
% }

\end{document}